\documentclass[10pt,twocolumn,letterpaper]{article}

\usepackage{booktabs}
\usepackage{multirow}
\usepackage{bbding}
\usepackage{makecell}
\usepackage{xcolor}
\usepackage{afterpage}
\usepackage{float}
\usepackage{placeins}
\usepackage{cuted}
\usepackage[pagenumbers]{cvpr} 

\definecolor{cvprblue}{rgb}{0.21,0.49,0.74}
\usepackage[pagebackref,breaklinks,colorlinks,allcolors=cvprblue]{hyperref}

\title{One-Step Diffusion Transformer for \\ Controllable Real-World Image Super-Resolution}

\author{
Yushun Fang$^{1,2,*}$ \quad 
Yuxiang Chen$^{2,*}$ \quad 
Shibo Yin$^{2,\dagger}$ \quad
Qiang Hu$^{1,\dagger}$ \quad
Jiangchao Yao$^{1}$ \\
Ya Zhang$^{1}$ \quad
Xiaoyun Zhang$^{1,\dagger}$ \quad
Yanfeng Wang$^{1}$ \\
$^{1}$Shanghai Jiao Tong University \quad $^{2}$Xiaohongshu Inc\\
$^{*}$Equal contribution \quad $^{\dagger}$Corresponding author
}

\begin{document}

\twocolumn[{%
\renewcommand\twocolumn[1][]{#1}%
\maketitle
\vspace{-2em}
\includegraphics[width=\linewidth]{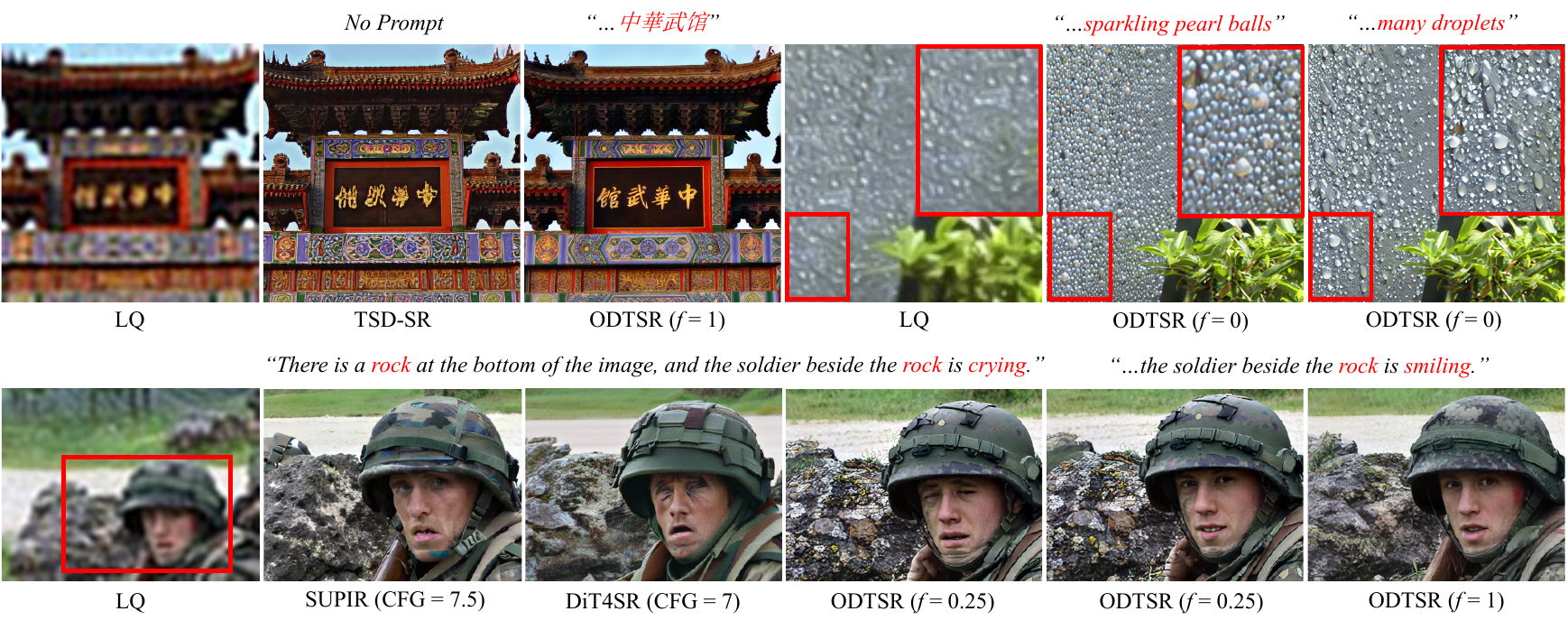}
\vspace{-2em}
\captionof{figure}{\textbf{Controllable Real-ISR:} Qualitative results of our ODTSR and other state-of-the-art methods. Our method achieves superior quality, supports flexible bilingual prompt controllability, and covers challenging sub-domains such as Chinese text images, fine-grained texture and face images. ``\textit{f}'' denotes controllable \textit{Fidelity Weight} in ODTSR. More results are shown in the \textbf{supplementary materials}.\vspace{1em}}
\label{fig:1.1}
}]

\begin{abstract}
Recent advances in diffusion-based real-world image super-resolution (Real-ISR) have demonstrated remarkable perceptual quality, yet the balance between fidelity and controllability remains a problem: multi-step diffusion-based methods suffer from generative diversity and randomness, resulting in low fidelity, while one-step methods lose control flexibility due to fidelity-specific finetuning. In this paper, we present \textbf{ODTSR}, a one-step diffusion transformer based on Qwen-Image that performs Real-ISR considering fidelity and controllability simultaneously: a newly introduced visual stream receives low-quality images (LQ) with adjustable noise (Control Noise), and the original visual stream receives LQs with consistent noise (Prior Noise), forming the Noise-hybrid Visual Stream (NVS) design. ODTSR further employs Fidelity-aware Adversarial Training (FAA) to enhance controllability and achieve one-step inference. Extensive experiments demonstrate that ODTSR not only achieves state-of-the-art (SOTA) performance on generic Real-ISR, but also enables prompt controllability on challenging scenarios such as real-world scene text image super-resolution (STISR) of Chinese characters without training on specific datasets. Codes are available at \url{https://github.com/RedMediaTech/ODTSR}.
\end{abstract}    
\section{Introduction}
\label{sec:intro}

\afterpage{
    \begin{table}[ht]
  \caption{\textbf{Overall Comparison.} We compare four SOTA methods with our proposed ODTSR. ODTSR achieves fidelity and prompt controllability simultaneously while achieving one-step inference. ``\CheckmarkBold\CheckmarkBold'' denotes ODTSR support both English and Chinese prompts.\vspace{-0.5em}}
  \label{Tab:intro}
  \centering
  \resizebox{0.85\columnwidth}{!}{
      \begin{tabular}{c|c c c}
        \toprule
        Method &  One step & Fidelity & Prompt Control \\
        \midrule
        PiSA-SR      & \CheckmarkBold & \CheckmarkBold  & \XSolidBrush \\
        TSD-SR    & \CheckmarkBold & \CheckmarkBold & \XSolidBrush \\
        SUPIR    & \XSolidBrush & \XSolidBrush & \CheckmarkBold \\
        DiT4SR    & \XSolidBrush & \CheckmarkBold & \CheckmarkBold  \\
        \midrule
        ODTSR (Ours)    & \CheckmarkBold & \CheckmarkBold & \CheckmarkBold\CheckmarkBold \\
        \bottomrule
      \end{tabular}
  }
  \vspace{-1em}
\end{table}

}

Real-ISR\cite{wang2021real} aims to restore real-world low-quality, low-resolution images (LQ) to corresponding high-quality, high-resolution images (HR). Fidelity is fundamental for Real-ISR, however, under conditions of complex degradation and ambiguous semantics of LQ, the restoration of HR is not unique, which constitutes a classic ill-posed problem, inherently leading to demands on controllability.

In recent years, diffusion models have achieved remarkable perceptual quality in Real-ISR. However, simultaneously ensuring fidelity and controllability remains a challenge in the field. Multi-step diffusion approaches possess powerful generative controllability through iterative denoising, but tend to suffer from randomness, resulting in divergence from LQ, as well as expensive inference time. In contrast, one-step methods offer efficiency advantages, yet often lose the inherent controllability of pretrained diffusion models due to extensive fine-tuning for high fidelity, making prompt controllability difficult across diverse scenarios.

To address these challenges, we propose ODTSR, a one-step diffusion model for Real-ISR based on Qwen-Image~\cite{wu2025qwen}, that jointly ensures fidelity and controllability. ODTSR introduces a Noise-hybrid Visual Stream (NVS) design in diffusion transformer~\cite{peebles2023scalable} (DiT) architecture: a new visual stream receives LQ with adjustable noise (Control Noise): when this noise level is minimal or even zero, a clean LQ ensures strong fidelity and the noise level can be increased as needed to enhance prompt controllability. Meanwhile, the original visual streams take in LQ images with consistent noise (Prior Noise) to inherit denoising capability from the pretrained diffusion model. For further efficiency and controllability, ODTSR adopts a Fidelity-aware Adversarial Training (FAA) strategy: fine-tuning pretrained DiT as the discriminator, adversarial signals are dynamically adjusted based on the level of Control Noise to enhance controllability and achieve one-step inference.

As shown in \cref{Tab:intro}, compared with current state-of-the-art (SOTA) methods, our proposed ODTSR jointly ensures fidelity and bilingual prompt controllability in a single-step manner. As a one-step method, ODTSR not only supports aiming at fidelity like other one-step SOTA methods PiSA-SR~\cite{sun2025pisasr} and TSD-SR~\cite{dong2025tsdsr}, but also enables prompt controllability like multi-step SOTA methods SUPIR~\cite{yu2024supir} and DiT4SR~\cite{duan2025dit4sr}. The proposed ODTSR outperforms all these methods in offering superior quality and greater flexibility as demonstrated by extensive experiments. ODTSR also shows strong generalization to cross-domain and long-tail scenarios; for instance, it achieves readable and style-consistent restoration of complex Chinese textual images without training on specific dataset, substantially reducing annotation and adaptation costs in practice.

To summarize, we make the following contributions:
\begin{itemize}[topsep=0pt,parsep=0pt,leftmargin=18pt]
    \item We introduce ODTSR, a one-step diffusion model for Real-ISR based on Qwen-Image~\cite{wu2025qwen}. To the best of our knowledge, this is the first one-step Real-ISR model with over 20B parameters that support prompts in both English and Chinese.
    \item We propose a Noise-hybrid Visual Stream (NVS) design, introducing a Control Noise stream for achieving fidelity and controllability simultaneously, while inheriting denoising capability from the pretrained diffusion model by denoising at Prior Noise stream. 
    \item We propose a Fidelity-aware Adversarial Training (FAA) paradigm adjusting adversarial signals dynamically to reinforce controllability and achieving one-step inference to sidestep the computational burden of multi-step sampling from a large diffusion model.
    \item Our proposed ODTSR achieves superior performance on generic Real-ISR datasets and supports flexible, bilingual prompt controllability, while covering challenging sub-domains such as Chinese text images, fine-grained texture and face images in the scene without specialty training as shown in \cref{fig:1.1}.
\end{itemize}
\section{Related Work}
\label{sec:related_work}

\paragraph{Diffusion-based Real-ISR.}
The success of large-scale text-to-image (T2I) diffusion models such as Stable Diffusion (SD)~\cite{rombach2022sd,podell2023sdxl,esser2024scaling} and Flux~\cite{flux2024,batifol2025flux} inspires researchers to leverage these pretrained models to address Real-ISR. StableSR~\cite{wang2024stablesr} incorporates LQs into UNet-based SD via trainable time-aware encoder. DiffBIR~\cite{lin2024diffbir} utilizes a ControlNet-like~\cite{zhang2023controlnet} reconstruction module. SeeSR~\cite{wu2024seesr} introduces degradation tags to guide the diffusion process. PASD~\cite{yang2024pasd} incorporates high-level semantic guidance to the diffusion model. OSEDiff~\cite{wu2024osediff} introduces VSD Loss~\cite{wang2023prolificdreamer,yin2024dmd} to achieve one-step inference. PiSA-SR~\cite{sun2025pisasr} proposes a dual-lora approach, considering pixel-level and semantic-level enhancement. TVT~\cite{yi2025tvt} trains a 4$\times$ VAE~\cite{kingma2013vae} module for SD to achieves better quality. Built upon SDXL~\cite{podell2023sdxl}, SUPIR~\cite{yu2024supir} introduces scaling-up in diffusion-based Real-ISR. When diffusion models enter the era of DiT~\cite{peebles2023scalable}, TSD-SR~\cite{dong2025tsdsr} first introduces target score
distillation to pretrained SD3~\cite{esser2024scaling} model. DiT4SR~\cite{duan2025dit4sr} investigates a more effective LQ conditioning mechanism in DiT compared to ControlNet-like module. FluxSR~\cite{li2025fluxsr} proposes flow trajectory distillation to distill a multi-step flow matching model into one-step. However, all the methods mentioned above cannot simultaneously achieve both fidelity and prompt controllability in a one-step manner, which leaves room for further exploration.
\section{Motivation}

\begin{figure*}[!ht]
    \centering
    \includegraphics[width=1\linewidth]{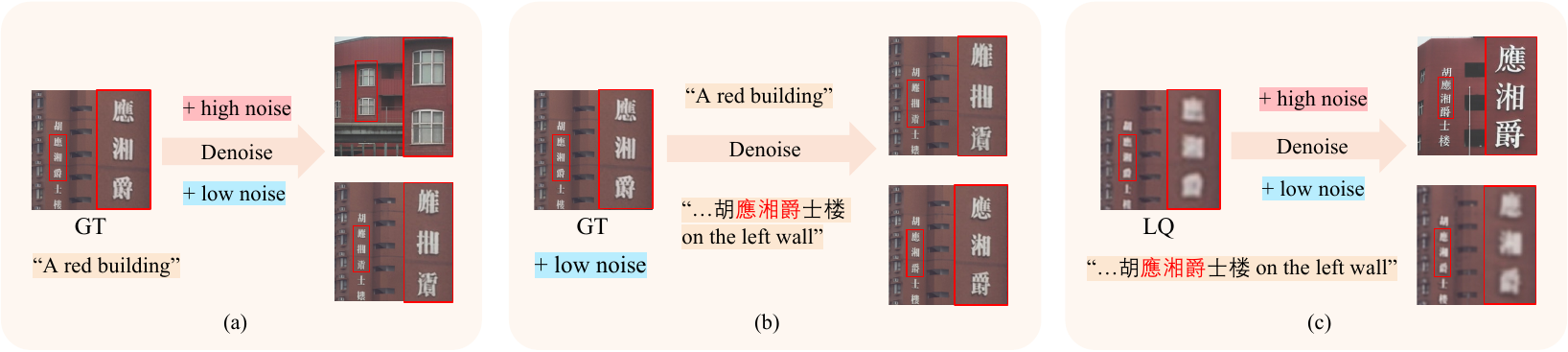}
    \vspace{-1.5em}
    \caption{\textbf{Effects of Noise on Fidelity and Controllability.} Based on pretrained Qwen-Image, \textit{(a)} shows the denoising results of ground-truth (GT) under the same prompt and different levels of noise. \textit{(b)} adopts low noise level and different prompts (with Chinese annotation). \textit{(c)} studies the effects of noise on LQ with the same prompt. High-noise inputs improve perceptual quality and controllability but reduce fidelity, whereas low-noise inputs preserve original details yet fail to deliver enhanced super-resolution effects. \vspace{-1em}}
    \label{fig:3.2}
\end{figure*}

\afterpage{%
    \begin{figure*}[!t]
        \centering
        \includegraphics[width=0.99\textwidth]{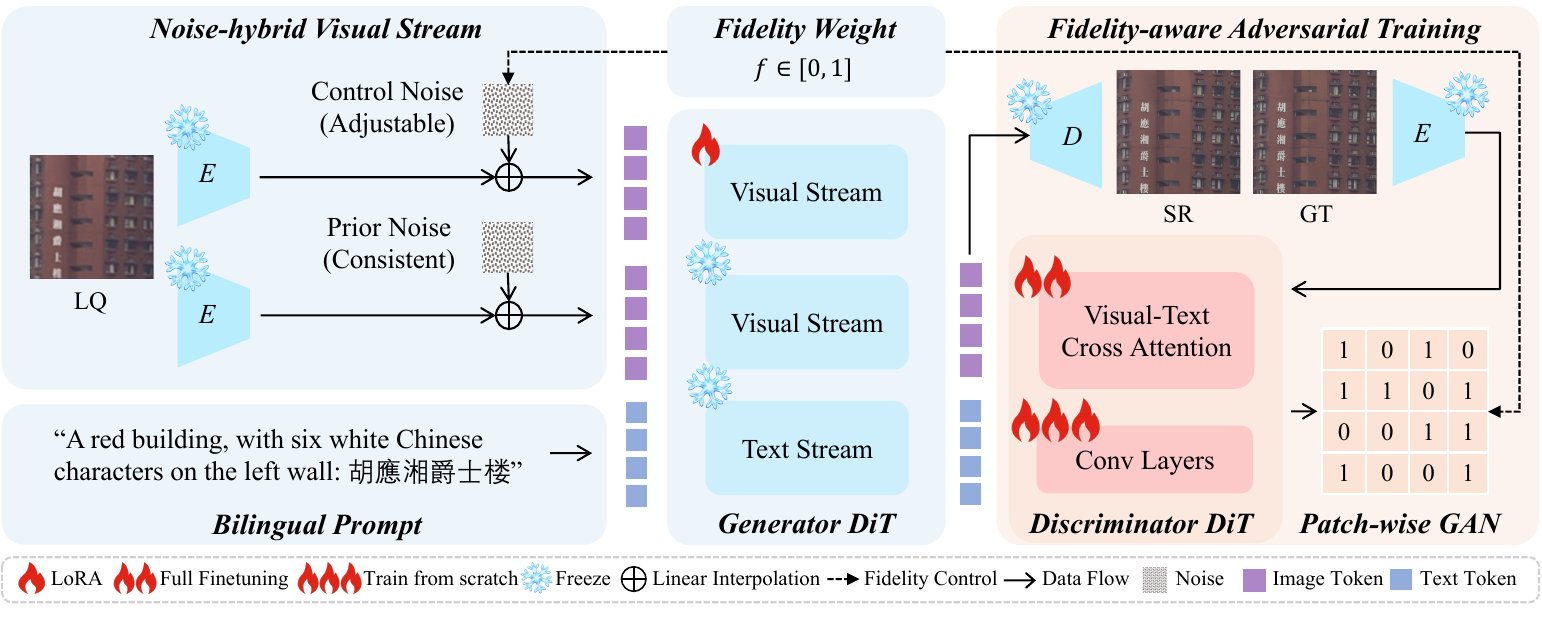}
        \caption{\textbf{Overall framework.} We extend the visual stream into a frozen \textit{Prior Noise} stream that preserves the pretrained diffusion prior and a \textit{Control Noise} stream that adaptively modulates noise based on the \textit{Fidelity Weight} to restore fine details, forming a \textit{Noise-hybrid Visual Stream}. Leveraging the \textit{Fidelity Weight} to adjust adversarial signals adaptively, \textit{Fidelity-aware Adversarial Training} enhances controllability and achieves one-step inference.\vspace{-1em}}
        \label{fig:3.3_main}
    \end{figure*}
}

In this section, we begin with an intuitive analysis of the Qwen-Image~\cite{wu2025qwen}, a leading-edge T2I model which demonstrates significant advances in text rendering and further supports prompts in both English and Chinese, demonstrating that such a model can inherently process images in a prompt-guided manner while achieving fidelity and controllability simultaneously through modifying the noise interpolation with the input image. However, LQs cannot be restored inherently only through this process, inspiring the proposed Noise-hybrid Visual Stream design.

\subsection{Preliminary}
Let $x_0\sim p_{\text{data}}$ denote a sample from the real data distribution, and let $x_1 \sim \mathcal{N}(0,I)$ be a sample from a simple prior (\eg Gaussian noise). Flow matching models define a continuous-time trajectory $x_t$ between $x_1$ and $x_0$, and learn a vector field $v_\theta(x,t)$ to match the data flow. According to Rectified Flow~\cite{liu2022flow}, the intermediate latent variable $x_t$ at timestep $t$ and its corresponding velocity $v_t$ can be calculated as:
\begin{equation}
x_t = t x_1 + (1-t) x_0, \quad
v_t = \frac{dx_t}{dt} = x_1 - x_0,
\label{eq:flow_matching}
\end{equation}
where $t \in [0,1]$. 
During training, the model minimizes the mean squared error loss:
\begin{equation}
\mathcal{L}(\theta) = \mathbb{E}_{(x_0,c) \sim \mathcal{D}}  \| v_\theta(x_t, t, c) - v_t \|^2 ,
\end{equation}
where $v_\theta(x_t, t, c)$ is the velocity predicted by the model, $c$ denotes the text prompt, and $\mathcal{D}$  denotes the training dataset.

\subsection{Effects of Noise on Fidelity and Controllability}
\label{sec:3.2}
In this section, we analyze the performance of Qwen-Image~\cite{wu2025qwen} when directly applied to Real-ISR tasks through noise addition and denoising, especially considering the fidelity and prompt controllability.

At intermediate timesteps during inference, the model naturally performs a dual-conditioned task: it predicts the velocity $v_\theta(x_t, t, c)$ based on the noised latent $x_t$ and text prompt $c$.
The inherent conditioning mechanism on partially noised inputs and prompt guidance enables controlled image manipulation. To investigate its applicability to Real-ISR, we begin by conducting a controlled experiment on a ground-truth (GT) high-resolution image $x_\text{GT}$, before extending the analysis to low-quality inputs. Specifically, we select two different timesteps $t_s \in \{0.90, 0.29\}$ as the starting points and construct corresponding noisy intermediates $x_{t_s}$ by linearly interpolate the clean image $x_\text{GT}$ and the noise $\epsilon \sim \mathcal{N}(0,I)$. 
According to \cref{eq:flow_matching}, the initialization at timestep $t$ is given by :
\begin{equation}
x_{t_s} = (1-t_s) x_\text{GT} + t_s \epsilon, \quad
\epsilon \sim \mathcal{N}(0,I),
\end{equation}
where larger $t_s$ values indicate stronger noise corruption and smaller $t_s$ values retain more information from the original image $x_\text{GT}$. 
Each noisy intermediate $x_{t_s}$ is then denoised through the remaining diffusion steps, conditioned on the input prompt $c$. 

This setup allows us to visualize how different noise levels affect the reconstruction fidelity and prompt controllability. As shown in \cref{fig:3.2}(a), under strong noise level (high $t$ values, ``\textit{+high noise}''), although most of the original structures are corrupted (e.g. Chinese characters turn into windows), the diffusion model is still able to recover semantic structures guided by the prompt, while maintaining a certain degree of low-level consistency and an acceptable image quality. Meanwhile, under light noise level (low $t$ values, ``\textit{+low noise}''), most of the structure is retained, while fine-grained structure like Chinese characters are corrupted. Then we further introduce specific text prompt as shown in \cref{fig:3.2}(b). Under light noise level, text prompt still works surprisingly as it fixes the corrupted Chinese characters while retaining the other original structures. 

We then replace the clean ground-truth sample $x_\text{GT}$ with a low-quality image $x_\text{LQ}$.  
The corresponding noisy intermediate is then constructed as:
\begin{equation}
x_{t_s} = (1-t_s) x_\text{LQ} + t_s \epsilon, \quad
\epsilon \sim \mathcal{N}(0,I).
\label{eq:xts}
\end{equation}
As shown in \cref{fig:3.2}(c), under strong noise level, although the diffusion model is able to significantly improve image quality and reconstruct Chinese characters guided by the prompt, low-level fidelity is unacceptably corrupted. Under light noise level, the image quality remains low, although the fine-grained structures (Chinese characters) are slightly clearer compared with LQ. 

In a nutshell, regardless of the quality of the input image, a higher noise level leads to stronger prompt controllability and higher quality, and low noise level maintains low-level fidelity while still having prompt-guided fine-grained control. This comes very close to ensuring both fidelity and prompt controllability in Real-ISR, and most importantly, it is the capability that the pretrained model possesses out-of-the-box without requiring any training. This inspires us to make optimal adaptations to the pretrained DiT model based on denoising capability tailored to varying noise levels. We thus propose the Noise-hybrid Visual Stream design.

\section{Method}

\afterpage{
    \begin{table*}[!ht]
  \caption{\textbf{Real-ISR:} Quantitative comparisons on \textit{RealSR}, \textit{DRealSR}, and \textit{DIV2K-Val}. The best and the second-best are highlighted in \textcolor{red}{\textbf{bold}} and \textcolor{blue}{\underline{underline}}. ``CFG'' denotes ``classifier free guidance scale'' in multi-step methods, and ``f'' denotes ``fidelity weight'' in ODTSR.\vspace{-0.5em}}
  \label{Tab:main}
  \centering
  \resizebox{0.92\textwidth}{!}{
      \begin{tabular}{c|c|c | c c | c c | c | c c}
        \toprule
        Dataset & Method & Step(s) & LPIPS$\downarrow$ & DISTS$\downarrow$ & MUSIQ$\uparrow$ & MANIQA$\uparrow$ & FID$\downarrow$ & PSNR$\uparrow$ & SSIM$\uparrow$ \\
        \midrule
        \multirow{7}{*}{RealSR} 
        & OSEDiff  & 1 & 0.2921 & 0.2128 & 69.09 & 0.6326 & 123.49 & 25.15 & 0.7341  \\
        & PiSA-SR     & 1 & 0.2672 & \textcolor{blue}{\underline{0.2044}} & \textcolor{blue}{\underline{70.15}} & 0.6560 & 124.09 & \textcolor{blue}{\underline{25.50}} & \textcolor{blue}{\underline{0.7417}}  \\
        & TSD-SR    & 1 & 0.2743 & 0.2104 & \textbf{\textcolor{red}{71.19}} & 0.6347 & 114.45 & 24.81 & 0.7172  \\
        & TVT    & 1 & \textcolor{blue}{\underline{0.2597}} & 0.2061 & 69.89 & 0.6232 & \textcolor{blue}{\underline{109.90}} & \textbf{\textcolor{red}{25.81}} & \textbf{\textcolor{red}{0.7596}}  \\
        & SUPIR (CFG=4.0)      & 50 & 0.3541 & 0.2488 & 62.09  & 0.5780 & 130.38 & 23.65 & 0.6620  \\
        & DiT4SR (CFG=8.0)     & 40 & 0.3215 & 0.2251 & 67.76  & \textcolor{blue}{\underline{0.6564}} & 118.55 & 23.50 & 0.6657  \\
        & ODTSR (f=1.0, w/o prompt) & 1 & \textbf{\textcolor{red}{0.2398}} & \textbf{\textcolor{red}{0.1894}} & 68.29  & \textbf{\textcolor{red}{0.6622}} & \textbf{\textcolor{red}{101.49}} & 25.07 & 0.7361 \\
        \midrule
        \multirow{7}{*}{DRealSR}
        & OSEDiff  & 1 & 0.2968 & 0.2165 & 64.65 & 0.5895 & 135.29 & 27.92 & \textcolor{blue}{\underline{0.7835}}  \\
        & PiSA-SR     & 1 & 0.2960 & 0.2169 & \textcolor{blue}{\underline{66.11}} & 0.6146 & \textcolor{blue}{\underline{130.48}} & \textbf{\textcolor{red}{28.32}} & 0.7804  \\
        & TSD-SR   & 1 & 0.2966 & \textcolor{blue}{\underline{0.2136}} & \textbf{\textcolor{red}{66.61}} & 0.5850 & 135.32 & 27.77 & 0.7559  \\
        & TVT  & 1 & \textcolor{blue}{\underline{0.2899}} & 0.2205 & 65.56 & 0.5776 & 134.28 & \textcolor{blue}{\underline{28.27}} & \textbf{\textcolor{red}{0.7899}}  \\
        & SUPIR (CFG=4.0) & 50 & 0.4243 & 0.2795 & 58.79 & 0.5471 & 169.48 & 25.09 & 0.6460  \\
        & DiT4SR (CFG=8.0) & 40 & 0.3869 & 0.2508 & 65.75 & \textbf{\textcolor{red}{0.6287}} & 163.05 & 25.40 & 0.6657  \\
        & ODTSR (f=1.0, w/o prompt)  & 1 & \textbf{\textcolor{red}{0.2592}} & \textbf{\textcolor{red}{0.1926}} & 62.86 & \textcolor{blue}{\underline{0.6227}} & \textbf{\textcolor{red}{119.86}} & 28.14 & 0.7736  \\
        \midrule
        \multirow{7}{*}{DIV2K-Val}
        & OSEDiff   & 1 & 0.2942 & 0.1975 & 67.96 & 0.6147 & 26.34 & 23.72 & \textcolor{blue}{\underline{0.6109}}  \\
        & PiSA-SR     & 1 & 0.2823 & 0.1934 & \textcolor{blue}{\underline{69.68}} & 0.6401 & 25.09 & 23.87 & 0.6058  \\
        & TSD-SR  & 1 & \textbf{\textcolor{red}{0.2673}} & \textcolor{blue}{\underline{0.1821}} & \textbf{\textcolor{red}{71.69}} & 0.6226 & \textcolor{blue}{\underline{23.29}} & 23.02 & 0.5808  \\
        & TVT   & 1 & 0.2773 & 0.1860 & 68.67 & 0.6060 & 24.78 & \textbf{\textcolor{red}{24.23}} & \textbf{\textcolor{red}{0.6292}}  \\
        & SUPIR (CFG=4.0) & 50 & 0.3919 & 0.2312 & 63.86 & 0.5903 & 31.40 & 22.13 & 0.5259  \\
        & DiT4SR (CFG=8.0)  & 40 & 0.3448 & 0.2100 & 68.07 & \textcolor{blue}{\underline{0.6430}} & 30.57 & 21.78 & 0.5485  \\
        & ODTSR (f=1.0, w/o prompt)  & 1 & \textcolor{blue}{\underline{0.2761}} & \textbf{\textcolor{red}{0.1700}} & 67.94 & \textbf{\textcolor{red}{0.6438}} & \textbf{\textcolor{red}{21.47}} & \textcolor{blue}{\underline{23.92}} & 0.6108  \\
        \midrule
      \end{tabular}
  }
  \vspace{-1em}
\end{table*}

}

In this section, we first formulate the SR task as a one-step denoising process, then introduce the Noise-hybrid Visual Stream (NVS) for achieving fidelity and prompt controllability simultaneously. To further enhance controllability, a Fidelity-aware Adversarial Training (FAA) scheme is proposed. Finally, we describe the training objectives of the generator and discriminator under adversarial training. The whole framework is shown in \cref{fig:3.3_main}.

\subsection{Model Formulation}
Let $E$ and $D$ denote the encoder and decoder of the pretrained variational auto-encoder (VAE), which map images to and from the latent space $z$. We denote by $I_\text{LQ}$ and $I_\text{GT}$ the low-quality and ground-truth RGB images, respectively, and by $x_\text{LQ} = E(I_\text{LQ})$ and $x_\text{GT} = E(I_\text{GT})$ their corresponding latent representations.
Our goal is to obtain the restored image $I_\text{SR} = D(x_\text{pred})$ from the predicted latent feature $x_\text{pred}$.

The Diffusion Transformer (DiT) architecture~\cite{peebles2023scalable} is a widely adopted backbone, featuring a simpler structure than UNet and better scalability. Building upon DiT, MMDiT~\cite{esser2024scaling} proposes an effective strategy for multi-modal information fusion by retaining modality-specific linear projections for computing Q, K, and V, while restricting cross-modal interaction to attention operator.

The base model we employ, Qwen-Image~\cite{wu2025qwen}, also adopts a similar double-stream design, consisting of a visual and a textual stream.

\subsection{Noise-hybrid Visual Stream}
From the observations in \cref{sec:3.2}, directly applying a noised LQ latent to the pretrained visual stream presents a trade-off: high-noise inputs improve perceptual quality but reduce fidelity, whereas low-noise inputs preserve original details yet fail to deliver enhanced super-resolution effects.
To address this issue, we extend the original visual stream into two branches: a \textbf{Prior Noise} stream and a \textbf{Control Noise} stream. 

The Prior Noise stream injects a fixed level of noise to $x_\text{LQ}$, denoted by a hyperparameter $t_p$, following \cref{eq:flow_matching}:
\begin{equation}
x_{t_p} = (1-t_p) x_\text{LQ} + t_p \epsilon, \quad
\epsilon \sim \mathcal{N}(0,I).
\end{equation}
This stream remains frozen to preserve the pretrained diffusion prior as much as possible.

The Control Noise stream, in contrast, dynamically modulates the noise level according to a user-controllable \textbf{Fidelity Weight} $f \in [0, 1]$.
When $f=0$, the noise level $t_c$ matches $t_p$, whereas $f=1$ corresponds to no added noise ($t_c=0$).  
Intermediate values are linearly mapped from $[0,1]$ to $[t_p,0]$, yielding a fidelity controllable latent representation:
\begin{equation}
x_{t_c} = (1-t_c) x_\text{LQ} + t_c \epsilon, \quad t_c = (1 - f) \cdot t_p.
\end{equation}
During training, the value of $f$ is uniformly sampled from the range $[0, 1]$, and $t_p$ is empirically set to $0.43$.
The parameters of the Control Noise stream are initialized from the Prior Noise stream and finetuned via LoRA~\cite{hu2022lora}.

Finally, all streams are jointly attended via the multi-modal DiT to predict the velocity field that updates the noisy latent toward the high-quality target in a single step:
\begin{equation}
x_\text{pred} = x_{t_p} + (0 - t_p) \, v_\theta(x_{t_p}, x_{t_c}, t_p, c),
\end{equation}
\begin{equation}
I_\text{pred} = D(x_\text{pred}),
\end{equation}
where \(c\) denotes the text prompt.  
This formulation maintains alignment with the pretrained diffusion prior via the Prior Noise stream, while the Control Noise stream delivers a controllable fidelity-conditioning signal and adaptively restores high-frequency details, together constituting the generator of ODTSR.

\afterpage{
    \begin{table*}[!ht]
  \caption{\textbf{Controllable Real-ISR:} Quantitative comparisons on \textit{RealSR} and \textit{RealCE-Val}. The best and the second-best are highlighted in \textcolor{red}{\textbf{bold}} and \textcolor{blue}{\underline{underline}}. ``*'' suggests TVT is the only method trained on \textit{RealCE} training set.\vspace{-0.5em}}
  \label{Tab:control}
  \centering
  \resizebox{\textwidth}{!}{
      \begin{tabular}{c|c|c c|c c| c | c c | c c }
        \toprule
        Dataset & Method & LPIPS$\downarrow$ & DISTS$\downarrow$ & MUSIQ$\uparrow$ & MANIQA$\uparrow$ & FID$\downarrow$ & CLIP-T $\uparrow$ & NED$\uparrow$ & PSNR$\uparrow$ & SSIM$\uparrow$ \\
        \midrule
        \multirow{5}{*}{\makecell{RealSR \\ (w/ prompt \\ by default)}}
        & SUPIR (CFG=4.0) & 0.3708 & 0.2459 & 61.52 & 0.5960 & 117.02 & 33.97 & - & 24.82 & 0.6647 \\
        & SUPIR (CFG=1.0) & 0.4113 & 0.2808 & 55.84 & 0.5262 & 125.13 & 30.66 & - & \textbf{\textcolor{red}{25.66}} & 0.6501 \\
        & DiT4SR (CFG=8.0) & 0.3315 & 0.2244 & 67.36 & 0.6612 & 121.73 & 33.49 & - & 23.18 & 0.6493 \\
        & DiT4SR (CFG=1.0) & 0.2715 & 0.1963 & 61.94 & 0.6047 & 110.25 & 31.57 & - & 24.90 & 0.7254 \\
        & ODTSR (f=0.2) & 0.3212 & 0.2211 & \textbf{\textcolor{red}{69.11}} & \textbf{\textcolor{red}{0.6824}} & 112.10 & \textbf{\textcolor{red}{34.56}} & - & 21.70 & 0.6146 \\
        & ODTSR (f=1.0)  & \textbf{\textcolor{red}{0.2310}} & \textbf{\textcolor{red}{0.1815}} & 68.22 & \textcolor{blue}{\underline{0.6680}} & \textbf{\textcolor{red}{91.12}} & \textcolor{blue}{\underline{34.01}} & - & \textcolor{blue}{\underline{25.09}} & \textbf{\textcolor{red}{0.7382}} \\
        & ODTSR (f=1.0, w/o prompt)  & \textcolor{blue}{\underline{0.2398}} & \textcolor{blue}{\underline{0.1894}} & \textcolor{blue}{\underline{68.29}} & 0.6622 & \textcolor{blue}{\underline{101.49}} & 32.37 & - & 25.07 & \textcolor{blue}{\underline{0.7361}} \\
        \midrule
        \multirow{8}{*}{\makecell{RealCE-val \\ (w/o prompt \\ by default)}}
        & OSEDiff     & 0.2967 & 0.2277 & 64.83 & 0.5863 & 86.21 & - & 0.6927 & 18.36 & 0.6800 \\
        & PiSA-SR        & \textbf{\textcolor{red}{0.2825}} & 0.2277 & 64.83 & 0.6090 & 82.58 & - & 0.7137 & \textcolor{blue}{\underline{19.48}} & 0.6923 \\
        & TSD-SR      & 0.2940 & 0.2396 & \textbf{\textcolor{red}{66.98}} & 0.6057 & 86.22 & - & 0.7205 & 19.00 & 0.6625 \\
        & TVT*     & 0.2935 & 0.2348 & 66.50 & 0.5980 & 83.85 & - & 0.7179 & 19.27 & \textbf{\textcolor{red}{0.6898}} \\
        & SUPIR (CFG=4.0)       & 0.4203 & 0.2867 & 51.99 & 0.5030 & 87.23 & - & 0.6877 & \textbf{\textcolor{red}{19.58}} & 0.5784 \\
        & DiT4SR (CFG=8.0)      & 0.3721 & 0.2662 & 66.09 & 0.6443 & 95.90 & - & 0.6794 & 17.66 & 0.6013 \\
        & ODTSR (f=1.0)       & 0.2843 & \textcolor{blue}{\underline{0.2114}} & 66.52 & \textcolor{blue}{\underline{0.6523}} & \textcolor{blue}{\underline{70.05}} & - & \textcolor{blue}{\underline{0.7609}} & 18.78 & 0.6863 \\
        & ODTSR (f=1.0, w/ prompt)       & \textcolor{blue}{\underline{0.2830}} & \textbf{\textcolor{red}{0.2097}} & \textcolor{blue}{\underline{66.78}} & \textbf{\textcolor{red}{0.6565}} & \textbf{\textcolor{red}{68.05}} & - & \textbf{\textcolor{red}{0.8475}} & 18.75 & \textcolor{blue}{\underline{0.6891}} \\
        \bottomrule
      \end{tabular}
  }
  \vspace{-1em}
\end{table*}

}

\subsection{Fidelity-aware Adversarial Training}
Although we have made the input as consistent as possible with the pretrained design (i.e., using noised latent features), converting the remaining multi-step denoising process into a single step remains a challenging training problem. Inspired by Diffusion Adversarial Post-Training (APT)~\cite{lin2025diffusion}, since our input is the noised LQ latent rather than pure noise, we skip the discrete-time consistency distillation\cite{song2023consistency, song2023improved} stage and directly perform single-stage adversarial training under Real-ISR task.

\paragraph{Training objective of generator.}
We train the generator with the reconstruction loss $\mathcal{L}_{\text{rec}}$ in RGB space and adversarial loss $\mathcal{L}_{\text{adv}}^G$ in latent space. We
set $\mathcal{L}_{\text{rec}}$ as the weighted sum of MSE loss and LPIPS loss:
\begin{equation}
\label{eq:loss_rec}
\mathcal{L}_{\text{rec}} = \mathcal{L}_{\text{MSE}}(I_{\text{pred}}, I_{\text{GT}}) + \lambda_1\mathcal{L}_{\text{LPIPS}}(I_{\text{pred}}, I_{\text{GT}}),
\end{equation}
where $\lambda_1$ is a weighting scalar.
As for $\mathcal{L}_{\text{adv}}$, inspired by R3GAN~\cite{huang2024gan}, we replace the non-saturating GAN loss\cite{goodfellow2014generative}
used in APT by a relativistic gan loss\cite{jolicoeur2018relativistic} to avoid the potential mode collapse problem\cite{kossale2022mode}:
\begin{equation}
\mathcal{L}_{\text{adv}}^G =  - \mathbb{E}_{x_r} [\log(1 - R_{}(x_r, x_f))] - \mathbb{E}_{x_f} [\log(R_{}(x_f, x_r))],
\end{equation}
where $x_f = {x}_\text{pred}$ and $x_r = x_\text{GT}$. $R()$ is formulated as:
\begin{equation}
R(x_r, x_f) = \sigma(D(x_r, t, c) - D(x_f, t, c)),
\end{equation}
where $\sigma$ is the sigmoid function, $t$ is the timestep for DiT based discriminator, $c$ denotes the text prompt and $D$ is the non-transformed patchified discriminator output.

\paragraph{Fidelity-aware Formulation} Conventional adversarial super-resolution methods typically train the generator with a fixed combination of reconstruction loss $\mathcal{L}_{\text{rec}}$ and adversarial loss $\mathcal{L}_{\text{adv}}^G$, regardless of the degradation of the low-quality input.
However, under our fidelity-aware formulation, the latent input $x_{t_c}$ is dynamically perturbed according to the controllable Fidelity Weight $f$, leading to varying information availability and distinct reconstruction difficulties.
To ensure consistency with this design, we modulate the balance between $\mathcal{L}_{\text{rec}}$ and $\mathcal{L}_{\text{adv}}^G$ by the fidelity weight $f$, which defines the overall generator objective:
\begin{equation}
\label{eq:loss_faa}
\mathcal{L}_{G} = \mathcal{L}_{\text{rec}} + \big(f \, \lambda_{\min} + (1 - f) \, \lambda_{\max}\big) \, \mathcal{L}_{\text{adv}}^G,
\end{equation}
where $\lambda_{\min}$ and $\lambda_{\max}$ denote the minimum and maximum weights of the adversarial loss, corresponding to high-fidelity ($f=1$) and low-fidelity ($f=0$) inputs, respectively.  
Intuitively, as the Fidelity Weight $f$ decreases (more degraded input), the adversarial weight increases, encouraging the generator to synthesize realistic details according to prompt input; conversely, as $f$ increases (higher-quality input), the adversarial contribution is reduced, favoring reconstruction fidelity. 

\afterpage{
    \begin{figure*}[!ht]
        \centering
        \includegraphics[width=\linewidth]{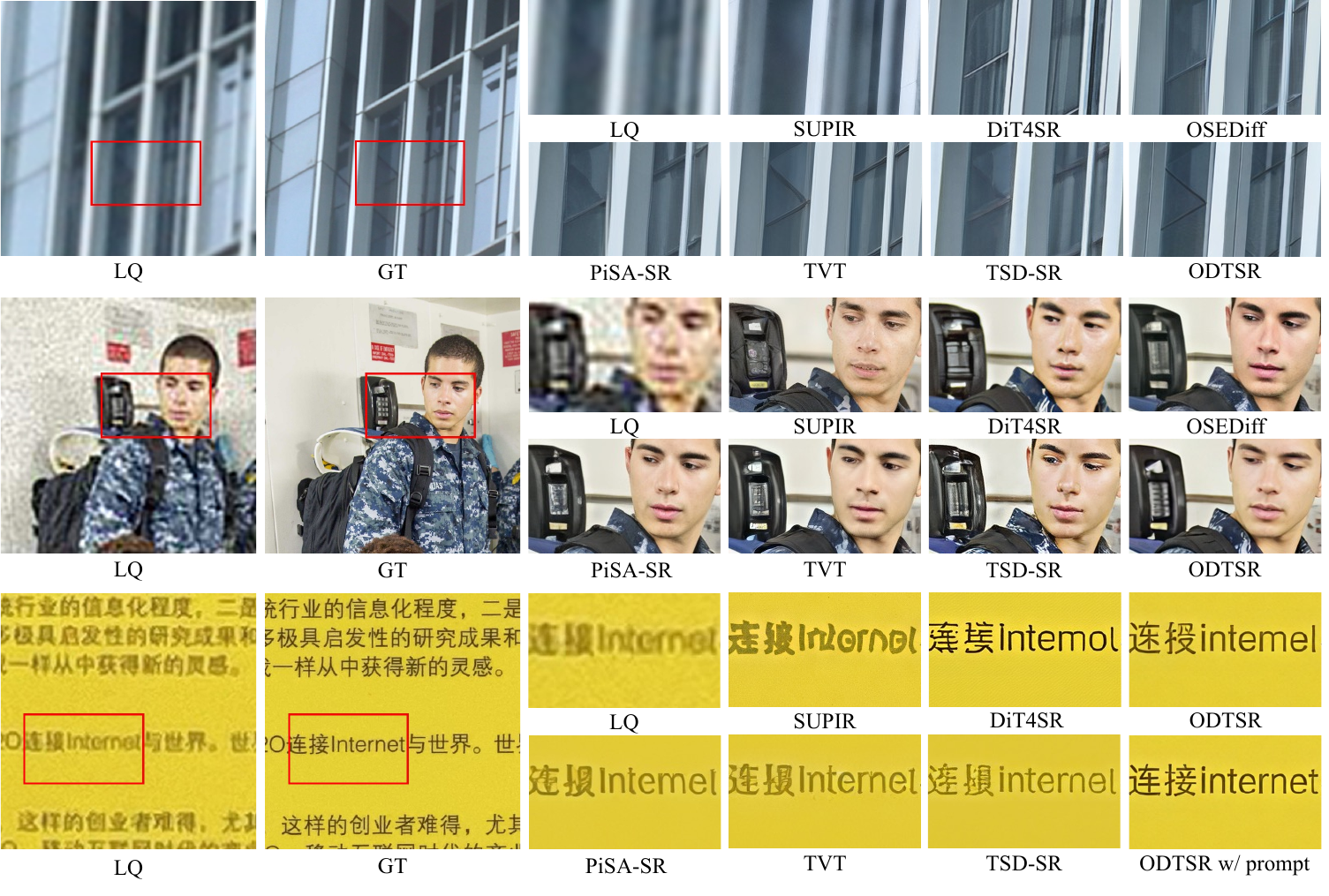}
        \vspace{-2em}
        \caption{\textbf{Real-ISR:} Qualitative comparison on texture (1st row), face image (2nd row) and text image (3rd row). ODTSR demonstrates superior performance in these challenging cases. More results are shown in the \textbf{supplementary materials}. \vspace{-1em}}
        \label{fig:exp_qualitative}
    \end{figure*}
}

\paragraph{Training objective of discriminator.}
We adopt the Wan2.1~\cite{wan2025wan} architecture as our discriminator, which leverages the same VAE encoder as Qwen-Image. Wan2.1 also builds upon the DiT backbone, in which textual conditioning is introduced through cross-attention. We initialize the discriminator with pretrained weights and fine-tune all parameters jointly. Unlike APT, we do not introduce additional cross-attention layers for producing a single scalar score. Instead, we append two 2D convolutional layers to the transformer outputs to produce patch-wise discrimination scores.

We train the discriminator with the adversarial loss $\mathcal{L}_{\text{adv}}^D$ and regularization loss $\mathcal{L}_{\text{reg}}$. The adversarial loss for the discriminator is formulated in a symmetrical form with respect to the generator loss $\mathcal{L}_{\text{adv}}^G$:
\begin{equation}
\mathcal{L}_{\text{adv}}^D =  - \mathbb{E}_{x_r} [\log( R_{}(x_r, x_f))] - \mathbb{E}_{x_f} [\log(1 - R_{}(x_f, x_r))].
\end{equation}
We also adopt the approximated R1 loss as regularization:
\begin{equation}
\mathcal{L}_{\text{reg}} = || D(x_r, t, c) - D(\mathcal{N}(x_r, \sigma I), t, c) ||_2^2,
\end{equation}
where $\mathcal{N}(x_r, \sigma I)$ denotes adding small-variance Gaussian perturbations to the real data $x_r$.
The loss encourages the discriminator to produce similar predictions for both the original and perturbed data, effectively reducing the discriminator’s gradient on real samples.
The overall training objective for the discriminator is:
\begin{equation}
\label{eq:loss_reg}
\mathcal{L}_{D} =  \mathcal{L}_{\text{adv}}^D + \lambda_2 \mathcal{L}_{\text{reg}},
\end{equation}
where $\lambda_2$ is a weighting scalar.
\section{Experiments}

\subsection{Experimental Settings}

\paragraph{Implementation.} We train ODTSR for the $\times$4 Real-ISR task based on Qwen-Image~\cite{wu2025qwen} as generator, and Wan2.1-T2V-1.3B~\cite{wan2025wan} as discriminator. The training process takes 10,000 iterations on 8 NVIDIA H20 GPUs with a batch size of 32. $\lambda_{1}=1.0$ in \cref{eq:loss_rec}. $\lambda_{\min}=0.02, \lambda_{\max}=0.1$ in \cref{eq:loss_faa}. $\lambda_{2}=5.0$ in \cref{eq:loss_reg}. Other details are shown in the \textbf{supplementary materials}.

\paragraph{Datasets.} For training, We utilize LSDIR~\cite{li2023lsdir} and FFHQ~\cite{karras2019style}, where LQs are synthesized online following common degradation pipeline proposed in Real-ESRGAN~\cite{wang2021real}. For Real-ISR evaluation, we adopt two real-world datasets: RealSR~\cite{cai2019realsr} and DRealSR~\cite{wei2020drealsr}, and a synthetic dataset DIV2K-Val~\cite{agustsson2017div2k} following OSEDiff~\cite{wu2024osediff}. In this scenario, our model's input prompt is left empty and other methods retain their default settings. For controllable Real-ISR evaluation, we adopt RealSR and a scene text image super-resolution (STISR) dataset RealCE's~\cite{ma2023realce} validation set (RealCE-val). Prompts of RealSR are extract from GT using Qwen2.5-VL-7B-Instruct~\cite{bai2025qwen25vl} and all methods adopt the same prompts. For RealCE-val, we center-crop all images to 512$\times$512 and use Paddle-OCR~\cite{cui2025paddleocr30technicalreport} to extract text as prompt. Notably, since ODTSR is the only method capable of processing Chinese prompts, we present results with and without prompts for comprehensive comparison. More details are in the \textbf{supplementary materials}.

\paragraph{Metrics.} Both full-reference (FR) and no-reference (NR) metrics are employed for evaluation.
FR metrics include pixel-level metrics PSNR and SSIM~\cite{wang2004image}, perceptual-level metrics LPIPS~\cite{zhang2018unreasonable} and DISTS~\cite{ding2020image} and distribution-level FID~\cite{heusel2017gans}. Besides, we adopt CLIP Score (CLIP-T) to measure prompt adherence in controllable Real-ISR using jina-clip-v2~\cite{koukounas2024jinaclip}. NR image quality metrics include IQA methods MUSIQ~\cite{ke2021musiq} and MANIQA~\cite{yang2022maniqa}. For STISR dataset RealCE-val, we additionally employ NED (normalized edit distance)~\cite{ma2023realce} to evaluate the text similarity between gound-truth (GT) and restored image (HR):
\begin{equation}
NED(P, G) = 1 - \frac{ED(P,G)}{\max(|P|,|G|)},
\label{eq:ned}
\end{equation}
where $P$ and $G$ denote text sequences extracted from HR and GT respectively, $ED(\cdot)$ denotes edit distance between two text sequences and $|\cdot|$ denotes the length of the text.

\paragraph{Compared Methods.} We compare ODTSR with SOTA diffusion-based one-step methods OSEDiff~\cite{wu2024osediff}, PiSA-SR~\cite{sun2025pisasr}, TSD-SR~\cite{dong2025tsdsr} and TVT~\cite{yi2025tvt}, and multi-step methods SUPIR~\cite{yu2024supir} and DiT4SR~\cite{duan2025dit4sr}. We obtain all results using their official open-source codes and models under their default settings unless otherwise specified.

\subsection{Comparison with State-of-the-Arts}

\paragraph{Quantitative Comparisons of Real-ISR}

\cref{Tab:main} compares the performance of our proposed ODTSR with other SOTA diffusion-based methods on three Real-ISR datasets. In general, ODTSR achieves superior results across perceptual and distribution fidelity metrics in terms of LPIPS, DISTS and FID, while demonstrating competitive performance in NR image quality metrics like MANIQA. We also present PSNR and SSIM results for reference, demonstrating the distinction between our approach and SOTA methods, as these metrics do not adequately reflect visual fidelity and quality.

\paragraph{Quantitative Comparisons of Controllable Real-ISR}

In prompt-guided RealSR, we conducted targeted comparisons with multi-step SOTA methods SUPIR and DiT4SR, evaluating restore quality and prompt adherence under adjustable control levels. Results are shown in \cref{Tab:control}. When aiming at fidelity (i.e. ``f=1.0'' in our method and low CFG scale in SUPIR and DiT4SR), ODTSR ensures leadership across all metrics including fidelity, image quality and prompt adherence, while SUPIR struggles to achieve better fidelity by adjusting CFG and DiT4SR meets a significant image quality drop in low CFG scale. When the restore target turns to prompt controllability (i.e. ``f=0.2'' in our method and high CFG scale in SUPIR and DiT4SR), ODTSR achieves the best prompt adherence and maintains relative high level image quality, while fidelity is reasonably reduced, which means our prompt controllability has greater and more flexible scope for adjustment.

In STISR dataset RealCE-val, ODTSR outperforms existing SOTA methods on text similarity NED by a remarkable margin even with no prompt, and with the addition of prompts, text restoration performance sees a further leap forward. It is important to emphasize that ODTSR was not trained on any specific scene text image dataset and its outstanding performance demonstrates that our proposed method effectively preserves the text rendering capability of the pretrained Qwen-Image.

\paragraph{Qualitative Comparisons}
\cref{fig:exp_qualitative} shows qualitative comparisons on Real-ISR and controllable Real-ISR. The first and second row show that ODTSR offers high quality restoration results in fine-grained textures and face images in severe degradations. Besides, the third row demonstrates ODTSR's capability of restoring English and Chinese text image, and the quality can be further boosted with the text annoation prompt guidance (denoted as ``ODTSR w/ prompt''), indicating the prompt controllability of ODTSR.

\subsection{Ablation Study}

\paragraph{Effectiveness of NVS} We validate the effectiveness of ODTSR's key design: Noise-hybird Visual Stream (NVS), or, to elaborate, why is it necessary to introduce an additional visual stream rather than directly adjust the noise on the original visual stream. We thus train a variant, denoted as ``1-visual'', which adds no additional visual stream and keeps all other training settings. Results are shown in \cref{Tab:ablation_structure}. When aiming at fidelity (``f=1.0''), regardless of whether prompt (extracted from GT) is involved, NVS shows notably better performance in fidelity, quality and prompt adherence. And in stronger prompt controllability setting (``f=0.2''), NVS sacrificing slightly lower fidelity (LPIPS and FID) for improved prompt adherence (CLIP-T) and image quality (MANIQA) means that even at the same noise level, NVS can better encourage prompt-guide generation, offering greater controllability.

\begin{table}[!ht]
  \caption{\textbf{Ablation of NVS.} We compare two variants of structure on \textit{RealSR} dataset. The better is highlighted in \textbf{bold}. ``p.'' denotes ``prompt''. \vspace{-0.5em}}
  \label{Tab:ablation_structure}
  \centering
  \resizebox{\columnwidth}{!}{
      \begin{tabular}{c|c c c c}
        \toprule
        Structure & LPIPS$\downarrow$ & MANIQA$\uparrow$ & FID$\downarrow$ & CLIP-T$\uparrow$ \\
        \midrule
        1-Visual (f=1.0, w/o p.)  & 0.2655 & 0.6387 & 118.08 & 32.01 \\
        NVS (f=1.0, w/o p.)  & \textbf{0.2398}  & \textbf{0.6622} & \textbf{101.49} & \textbf{32.37} \\
        \midrule
        1-Visual (f=1.0, w/ p.)  & 0.2552 & 0.6499 & 105.94 & 33.94 \\
        NVS (f=1.0, w/ p.)  & \textbf{0.2310} & \textbf{0.6622} & \textbf{101.49} & \textbf{34.01} \\
        \midrule
        1-Visual (f=0.2, w/ p.)  & \textbf{0.2915} & 0.6688 & \textbf{102.03} & 34.35 \\
        NVS (f=0.2, w/ p.)  & 0.3212 & \textbf{0.6824} & 112.10 & \textbf{34.56} \\
        \bottomrule
      \end{tabular}
  }
  \vspace{-1.5em}
\end{table}

\paragraph{Effectiveness of FAA} \cref{Tab:ablation_nat} demonstrates the effectiveness of Fidelity-aware Adversarial Training (FAA). Utilizing a fixed GAN weight, no matter low (0.02) or high (0.1), does not work well with the Control Noise stream as prompt adherence (CLIP-T) remarkably declines, since the model cannot learn to adjust the effect of prompt adaptively when Control Noise changes, and Control Noise alone corrupts the fidelity.

\begin{table}[!ht]
  \caption{\textbf{Ablation of FAA.} We compare two variants of GAN weight strategy on \textit{RealSR} (with prompt). As fidelity weight (``f'') is reduced from 1.0 to 0.2, performance metrics showing decline are marked in \textbf{bold}.}
  \label{Tab:ablation_nat}
  \centering
  \resizebox{\columnwidth}{!}{
      \begin{tabular}{c|c c| c c | c c}
        \toprule
        \makecell{GAN\\weight} & \makecell{FAA\\(f=1.0)} & \makecell{FAA\\(f=0.2)} & \makecell{0.02\\(f=1.0)} & \makecell{0.02\\(f=0.2)} & \makecell{0.1\\(f=1.0)} & \makecell{0.1\\(f=0.2)} \\
        \midrule
        CLIP-T & 34.01 & 34.56$\uparrow$ & 33.67 & \textbf{33.09$\downarrow$} & 33.92 & \textbf{32.71$\downarrow$} \\
        MANIQA & 0.668 & 0.682$\uparrow$ & 0.646 & 0.673$\uparrow$ & 0.636 & 0.671$\uparrow$ \\
        \bottomrule
      \end{tabular}
  }
  \vspace{-1.5em}
\end{table}

\paragraph{Ablation of Prior Noise level and other studies\hspace{-0.5em}}are also performed for a better understanding of ODTSR, which can be found in the \textbf{supplementary materials}.

\subsection{User Study}
To further demonstrate the performance, we conduct a user study, inviting 20 volunteers to evaluate ODTSR and other three latest SOTA methods. The results are presented in \cref{Tab:userstudy}, demonstrating that ODTSR's performance is more aligned with human preferences. More details can be found in the \textbf{supplementary materials}.
\begin{table}[!ht]
  \caption{\textbf{User study.} The better is highlighted in \textbf{bold}. Compared with other three SOTA methods, our model wins more than 50\% votes. \vspace{-0.5em}}
  \label{Tab:userstudy}
  \centering
  \resizebox{0.8\columnwidth}{!}{
      \begin{tabular}{c|c c c c}
        \toprule
        Method & Ours & TSD-SR & DiT4SR & PiSA-SR \\
        \midrule
        Votes  & \textbf{53.25\%} & 19.4\% & 12.75\% & 14.6\% \\
        \bottomrule
      \end{tabular}
  }
  \vspace{-1.5em}
\end{table}

\section{Conclusion}
In this work, we propose ODTSR, a one-step DiT that achieves fidelity and controllability via novel Noise-hybrid Visual Stream design and Fidelity-aware Adversarial Training strategy. We hope this work will advance the development of Real-ISR.

{
    \small
    \bibliographystyle{ieeenat_fullname}
    \bibliography{main}
}

\clearpage
\setcounter{page}{1}
\maketitlesupplementary

\section{Overview}
In this material, we provide the following contents:
\begin{itemize}
    \item More implementation details of ODTSR, including hyperparameters, more explanation of the Prior Noise stream, model structure and loss visualization (referring to Sec. 5.1 in the main paper);
    \item More details of training and testing dataset (referring to Sec. 5.1 in the main paper);
    \item More ablation studies related to the Prior Noise and GAN training strategies (referring to Sec. 5.3 in the main paper);
    \item User Study details (referring to Sec. 5.4 in the main paper);
    \item More qualitative comparison results of Real-ISR and controllable Real-ISR (referring to Sec 5.2 in the main paper);
    \item Limitation and future works (referring to Sec. 6 in the main paper)
\end{itemize}

\section{Implementation Details}
In this section, we present various details of the model to help readers gain a deeper understanding of its design.
\subsection{Hyperparameter Overview}
\begin{table}[ht]
  \vspace{-1.0em}
  \caption{A summary of the main hyperparameters used in ODTSR.\vspace{-0.5em}}
  \label{Tab:hyper}
  \centering
  \resizebox{\columnwidth}{!}{
      \begin{tabular}{c|c}
        \toprule
        \textbf{Hyperparameter} & \textbf{Value} \\
        \midrule
        Generator's Learning Rate & constant 5e-5 \\
        Discriminator's Learning Rate & constant 5e-6 \\
        Batch Size & 32 (gradient accumulation = 4)\\
        Generator's Optimizer & RMSprop (alpha=0.9, momentum=0.0) \\
        Discriminator's Optimizer & RMSprop (alpha=0.9, momentum=0.0) \\
        Lora Rank & 128 \\
        Lora Alpha & 128 \\
        Training Iters & 10,000 \\
        Training Resources & 8*NVIDIA H20 GPUs, within 48 hours \\
        $t_p$ in Eq.(5) of main paper & 0.43 \\
        Fidelity Weight $f$ & uniformly in the interval [0, 1] \\
                $\lambda_{1}$ in Eq.(9) of main paper & 1.0 \\
                $\lambda_\text{min}$ in Eq.(12) of main paper & 0.02 \\
                $\lambda_\text{max}$ in Eq.(12) of main paper & 0.1 \\
                variance $\sigma$ in Eq.(14) of main paper & 0.005 \\
                $\lambda_{2}$ in Eq.(15) of main paper & 5.0 \\
                
        Mixed Precision Training & True (\textit{float8\_e4m3fn} and \textit{bfloat16})  \\
        Gradient Checkpointing & True \\
        \bottomrule
      \end{tabular}
  }
  \vspace{-1em}
\end{table}

\begin{figure}[!ht]
    \centering
    \includegraphics[width=\columnwidth]{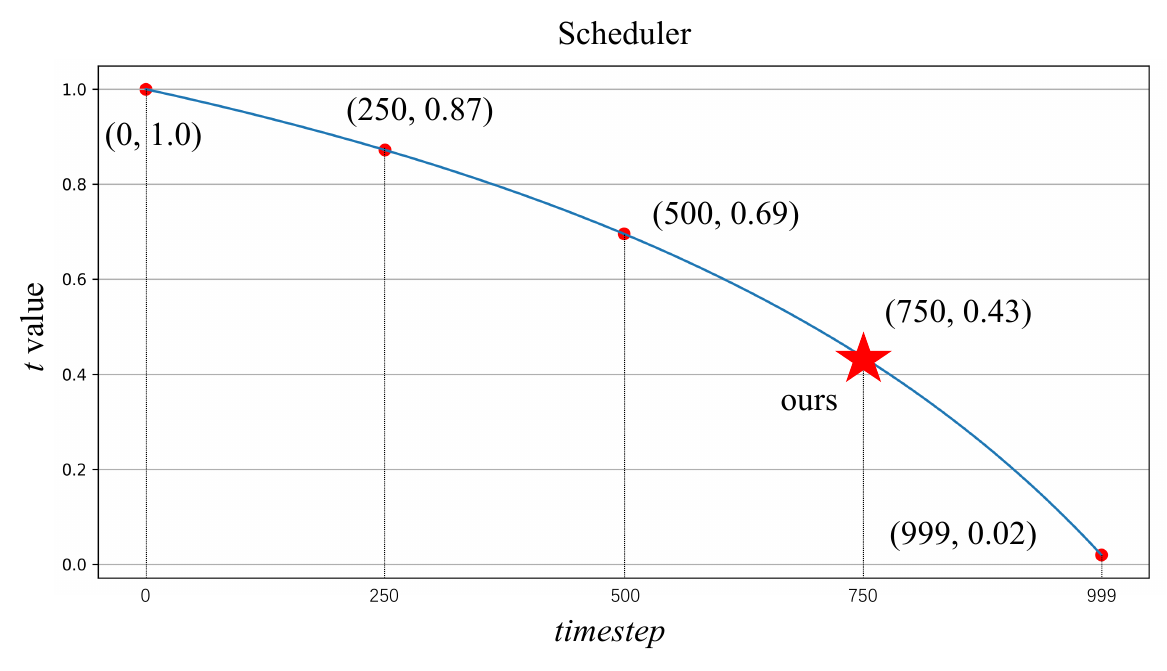}
    \caption{The \textbf{scheduler} used in Qwen-Image. The horizontal axis is the \textit{timestep} ranging from 0 to 999, 1000 discrete timesteps in total, and the vertical axis represents the values of $t$. During pre-training, \textit{timestep} is uniformly sampled to obtain $t$.}
    \label{fig:supp_curve}
\end{figure}

\afterpage{
    \begin{figure*}[!t]
        \centering
        \includegraphics[width=\linewidth]{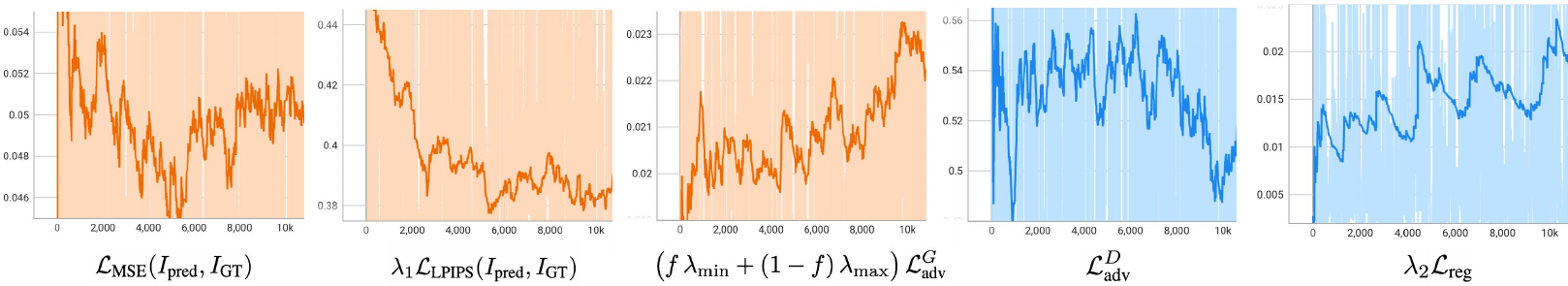}
        \caption{The visualization of the loss curves: the generator and discriminator losses are marked in different colors. Our training only needs 10,000 iterations to achieve good results.The meaning of the symbols corresponds to that in the main paper.}
        \label{fig:loss}
        \vspace{-1em}
    \end{figure*}
}

\subsection{Explanation of $t_p$ for Prior Noise stream}
In Eq.(5) of the main paper, the hyperparameter $t_p$ is a crucial value, as it determines the range of the noise level for LQ interpolation when the \textit{Fidelity Weight} changes. In T2I pretrained models based on Rectified Flow~\cite{liu2022flow}, a \textbf{scheduler} is typically used during pre-training. This scheduler establishes a relationship between \textit{timestep} (ranging from 0 to 999, 1000 discrete timesteps in total) and $t$ in Eq.(1) of the main paper. By sampling the timestep, the corresponding $t$ can be obtained. The scheduler used in Qwen-Image~\cite{wu2025qwen} is illustrated in \cref{fig:supp_curve}.

In our single-step training, we determine the value of $t_p$ by deciding the timestep, which is aligned with pre-training. This is why the value 0.43 looks a bit unusual, which actually corresponds to timestep 750.

We also ablate different timestep (i.e. 250 and 500) and use their corresponding $t_p$ values (i.e. 0.87 and 0.69) to train the model. See details in \cref{Sec:start timestep}.

\subsection{Model Structure Visualization}
\begin{figure*}[p]
    \centering
    \includegraphics[width=\linewidth]{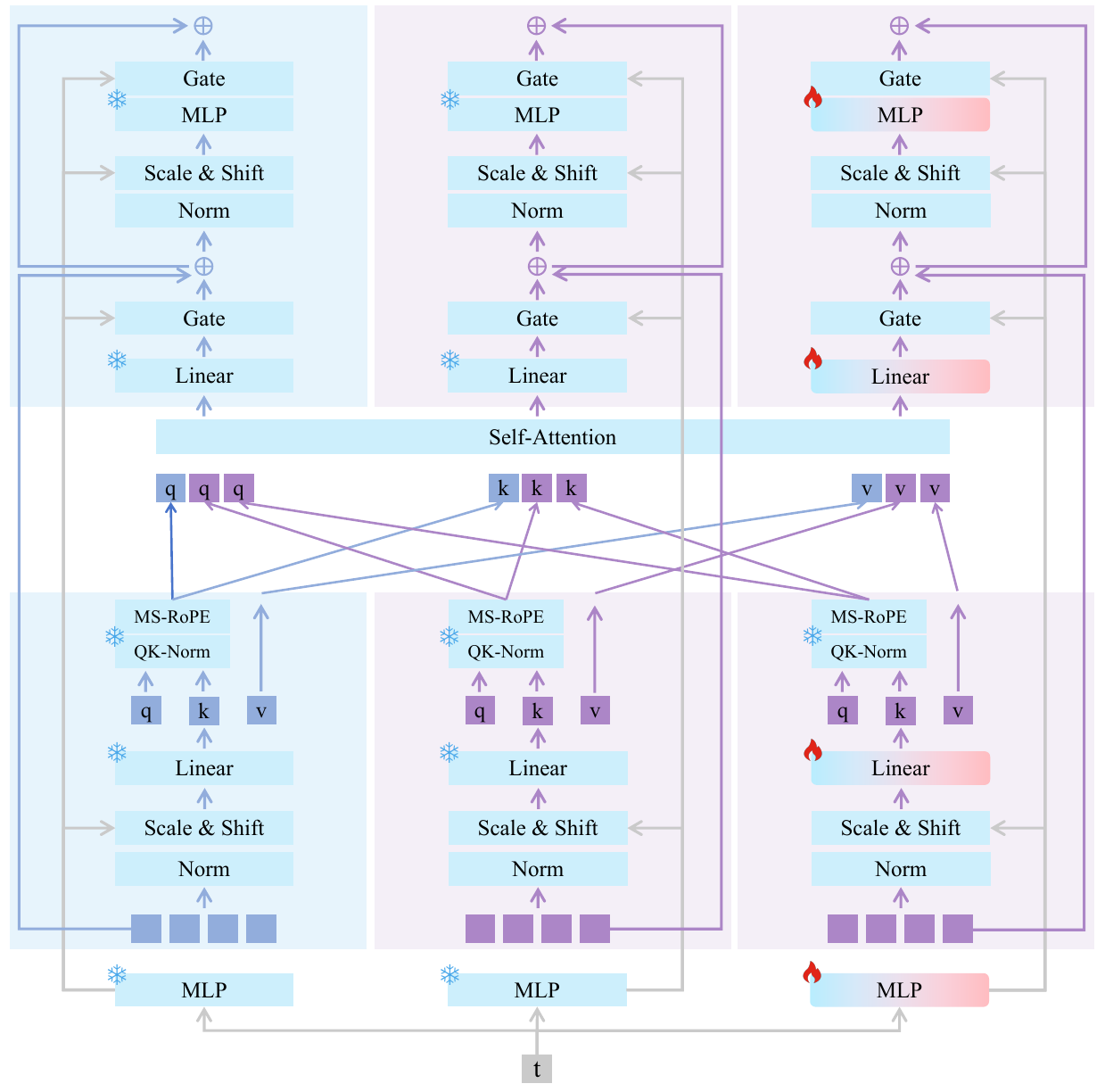}
    \caption{The model contains 60 transformer layers in total. The details of a single transformer layer are shown here: the left branch corresponds to the Text stream, the middle to the Prior Noise stream, and the right to the Control Noise stream. Among them, only the linear layers in the Control Noise stream are trained with LoRA, while all other parameters remain frozen.}
    \label{fig:supp_one_layer}
\end{figure*}

The base model we employ, Qwen-Image~\cite{wu2025qwen},  adopts a double-stream design, consisting of a visual and a textual stream. We provide the detailed structure of a single transformer layer, as shown in \cref{fig:supp_one_layer}.

\subsection{Loss Curve Visualization}
To help readers better understand the training dynamics, we visualize the loss curves, as shown in \cref{fig:loss}. Moreover, we summarize the common patterns observed in stable convergence as follows:

\begin{itemize}
    \setlength{\itemsep}{3pt}
    \item \textbf{$\mathcal{L}_{\text{MSE}}(I_{\text{pred}}, I_{\text{GT}})$}: the loss decreases rapidly at the beginning and later fluctuates within a stable range.
    \item \textbf{$\mathcal{L}_{\text{LPIPS}}(I_{\text{pred}}, I_{\text{GT}})$}: shows a steady decrease (at least during the first 10k iterations).
    \item \textbf{$\mathcal{L}_{\text{adv}}^G$}: shows a slight upward trend amid fluctuations.
    \item \textbf{$\mathcal{L}_{\text{adv}}^D$}: shows a slight downward trend amid fluctuations, and the value is less than $-\ln(\text{sigmoid}(0)) = -\ln(0.5) \approx 0.693$. This indicates that, on average, the discriminator has the ability to distinguish between real and fake samples.

    \item \textbf{$\mathcal{L}_{\text{reg}}$}: shows a slight upward trend amid fluctuations.
\end{itemize}

\afterpage{
\begin{figure*}[!ht]
    \centering
    \includegraphics[width=0.95\linewidth]{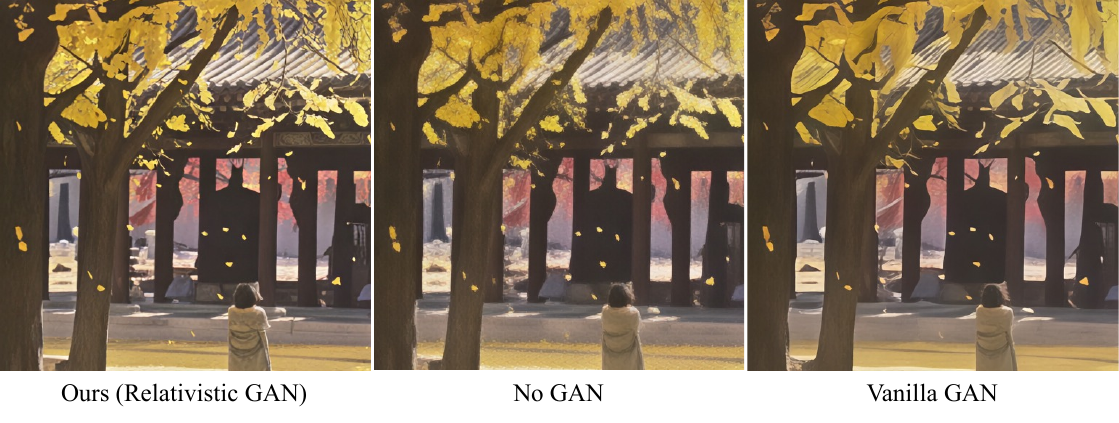}
    \caption{Visualization of GAN ablation. Trained with no GAN, model generates grid-like and blur artifacts. Trained with Vanilla GAN rather than Relativistic GAN, the restored image lacks detail and has been smoothed out.}
    \label{fig:supp_abla_gan}
    \vspace{-1em}
\end{figure*}
}

\section{Dataset Details}
\label{Sec:dataset}
\subsection{Training Set}
Like most previous methods, we use the LSDIR~\cite{li2023lsdir} and the first 10{,}000 images of FFHQ~\cite{karras2019style} as our training sets. The paired 512x512 data are constructed using the standard Real-ESRGAN~\cite{wang2021real} degradation pipeline. Specifically, the degradation hyperparameter configurations are exactly the same as those in PiSA-SR~\cite{sun2025pisasr}.

The difference is that our model requires textual descriptions during training. The discriminator always receives the description, while the generator uses the description with a probability of 0.75 and an empty description with a probability of 0.25 during each training iter.
We design a set of six prompt templates to extract textual descriptions from GT images using Qwen2.5-VL-7B-Instruct~\cite{bai2025qwen25vl}. These templates cover different levels of granularity, ranging from short tags to detailed captions and quality assessments:

\begin{itemize}
    \item "Use a few word tags to summarize the main content of this image."
    \item "Describe the main content of this image in one sentence."
    \item "Describe the content of this image in a few sentences."
    \item "Describe possible detailed features in this image, such as text or faces; if no such targets exist, describe other details; if multiple targets exist, describe them separately."
    \item "Describe the quality of this image and evaluate it based on factors such as clarity, color, noise, lighting, and focal length/lens effects."
    \item "Describe the artistic style or form of this image (e.g., photography, illustration, painting, CG) and explain the overall tone and mood."
\end{itemize}

Since the base model we use supports both English and Chinese, we include six additional Chinese templates, each corresponding to one of the English templates. During each training iteration, we randomly select one prompt from the total of 12 templates.

\subsection{Testing Set}


\paragraph{Controllable Real-ISR: RealSR}
We utilize the RealSR~\cite{cai2019realsr} dataset to evaluate the performance on general controllable image restoration.
The difference from a standard Real-ISR task is that we provide the model with a text description. The textual descriptions for RealSR are extracted from the ground truth images using Qwen2.5-VL-7B-Instruct~\cite{bai2025qwen25vl}, and all methods use the same descriptions.
The prompt used to extract the description is: "Describe the content of this image in a few sentences."
\paragraph{Controllable Real-ISR: RealCE}
We utilize the RealCE-val~\cite{ma2023realce} dataset to evaluate the performance on scene text image restoration, especially with text annotations. We center-crop all images to 512$\times$512 and manually remove pairs where LQ and GT are not aligned. Finally, we get 260 LQ-GT pairs. For text annotations, we employ \textit{PP-OCRv5} of Paddle-OCR~\cite{cui2025paddleocr30technicalreport} to extract text from GT and use it directly as the corresponding prompt.

\section{More Ablation Studies}
\subsection{Selection of timestep in the Prior Noise stream}
In this section, we investigate the impact of different \textit{timestep} selections of Prior Noise. We train another two models with different timestep (i.e. 250 and 500) for the same training steps and compare their performance. As shown in \cref{Tab:abl_prior_noise}, as timestep decreases from 750 to 250. which means $t_p$ and the level of the Prior Noise increases, the reconstruction fidelity is severely compromised. This is because our Fidelity-Aware Adversarial Training strategy samples GAN weight based on $t_p$, and higher $t_p$ value requires more training for fidelity (i.e. $f$ approaches 1 in Eq.(6) and Eq.(12) of the main paper). We thus choose timestep 750 empirically.

\label{Sec:start timestep}
\begin{table}[!h]
  \caption{\textbf{Ablation of timestep of Prior Noise.} We train two models with different timestep (i.e. 250 and 500) of Prior Noise for the same training steps and compare the performance on \textit{RealSR} dataset. \textit{Fidelity Weight} is set to 1.0. The best is highlighted in \textbf{bold}. \vspace{-0.5em}}
  \label{Tab:abl_prior_noise}
  \centering
  \resizebox{\columnwidth}{!}{
      \begin{tabular}{c | c |c c c c}
        \toprule
        Timestep & $t_p$  &LPIPS$\downarrow$ & MANIQA$\uparrow$ & FID$\downarrow$ & CLIP-T$\uparrow$ \\
        \midrule
        250  & 0.87 & 0.3318 & 0.6610 & 144.79 & 29.76 \\
        500  & 0.69 & 0.3057 & 0.6387 & 118.08 & 31.00 \\
        750 (Ours) & 0.43 & \textbf{0.2398}  & \textbf{0.6622} & \textbf{101.49} & \textbf{32.37} \\
        \bottomrule
      \end{tabular}
  }
\vspace{-1.5em}
\end{table}

\subsection{GAN training strategies}
In this section, we investigate two key questions in GAN training: (1) why GAN is necessary and (2) why relativistic GAN~\cite{jolicoeur2018relativistic} is chosen. We train two models following these two questions and evaluate their performance quantitatively and qualitatively. As shown in \cref{Tab:abl_gan} and \cref{fig:supp_abla_gan}, trained without GAN (``No GAN") is comparable in terms of fidelity, but fails in image quality, which is also demonstrated in visualization, showing grid-like artifacts and blur; trained with vanilla GAN (``Vanilla GAN") solves grid-like artifacts but produces over-smooth restoration results and also falls behind in the metric evaluation. Overall, relativistic GAN demonstrates the best performance. This also demonstrates that PSNR cannot accurately reflect image quality.

\begin{table}[!h]
  \caption{\textbf{Ablation of GAN Strategies.} \vspace{-0.5em}}
  \label{Tab:abl_gan}
  \centering
  \resizebox{\columnwidth}{!}{
      \begin{tabular}{c |c c c c}
        \toprule
        GAN  & LPIPS$\downarrow$ & MANIQA$\uparrow$ & FID$\downarrow$ & PSNR$\uparrow$ \\
        \midrule
        No GAN  & 0.2600 & 0.6180 & 119.28 & \textbf{26.07} \\
        Vanilla GAN  & 0.2931 & 0.6451 & 142.93 & 25.60 \\
        Ours (Relativistic GAN) & \textbf{0.2398}  & \textbf{0.6622} & \textbf{101.49} & 25.07 \\
        \bottomrule
      \end{tabular}
  }
  \vspace{-1.5em}
\end{table}

\section{User Study Details}
\begin{figure}[!h]
    \centering
    \includegraphics[width=\columnwidth]{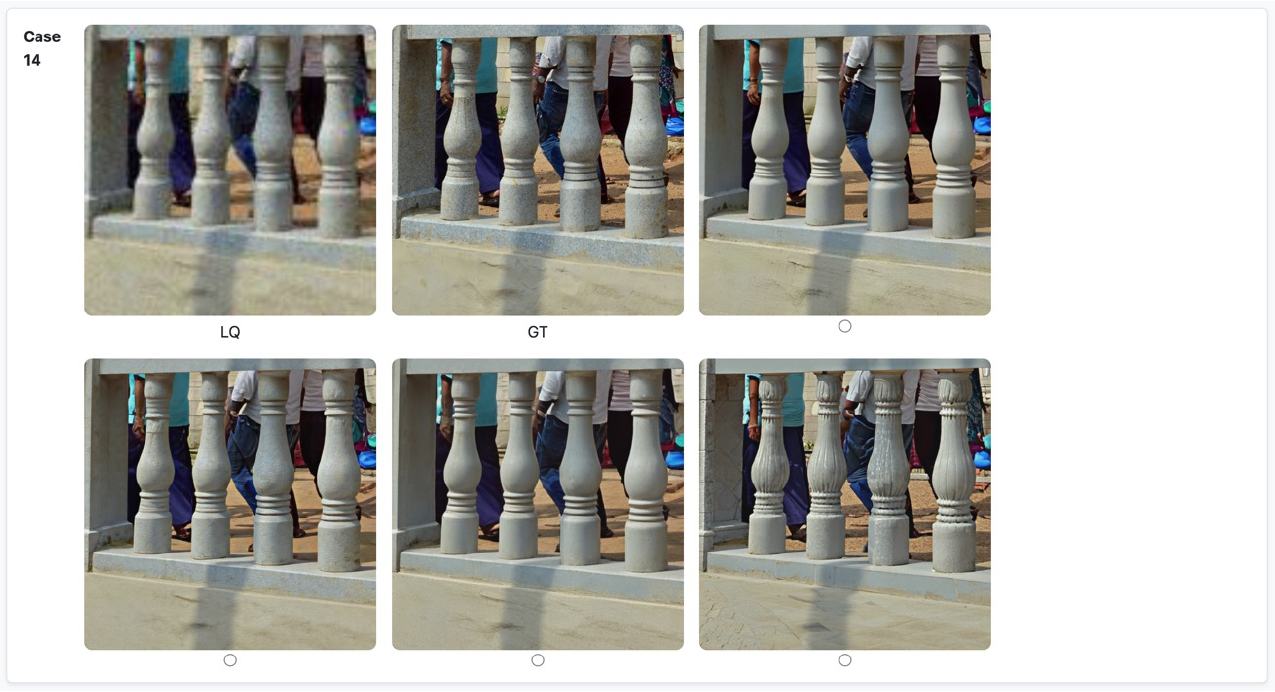}
    \caption{User Study Interface.}
    \label{fig:user_study_inferface}
    \vspace{-1em}
\end{figure}

We invite 20 volunteers to evaluate ODTSR and other three latest SOTA methods (TSD-SR~\cite{dong2025tsdsr}, PiSA-SR~\cite{sun2025pisasr} and DiT4SR~\cite{duan2025dit4sr}) in generic Real-ISR setting. 100 LQ images are randomly chosen from four datasets (RealSR~\cite{cai2019realsr}, DRealSR~\cite{wei2020drealsr}, Div2k-val~\cite{agustsson2017div2k}, RealCE-val~\cite{ma2023realce}). \textit{Fidelity Weight} is set to 1.0 and input prompt is empty in ODTSR. In the user study, volunteers are presented with six images: LQ, GT and restoration results from four methods randomly ordered. Volunteers are asked to choose the best result according to (1) fidelity with LQ and GT (2) overall image quality. We build a simple webpage for the user study, with the interface shown in \cref{fig:user_study_inferface}.

\section{More Qualitative Results}
\subsection{Real-ISR}
\begin{figure*}[p]
    \centering
    \includegraphics[width=\linewidth]{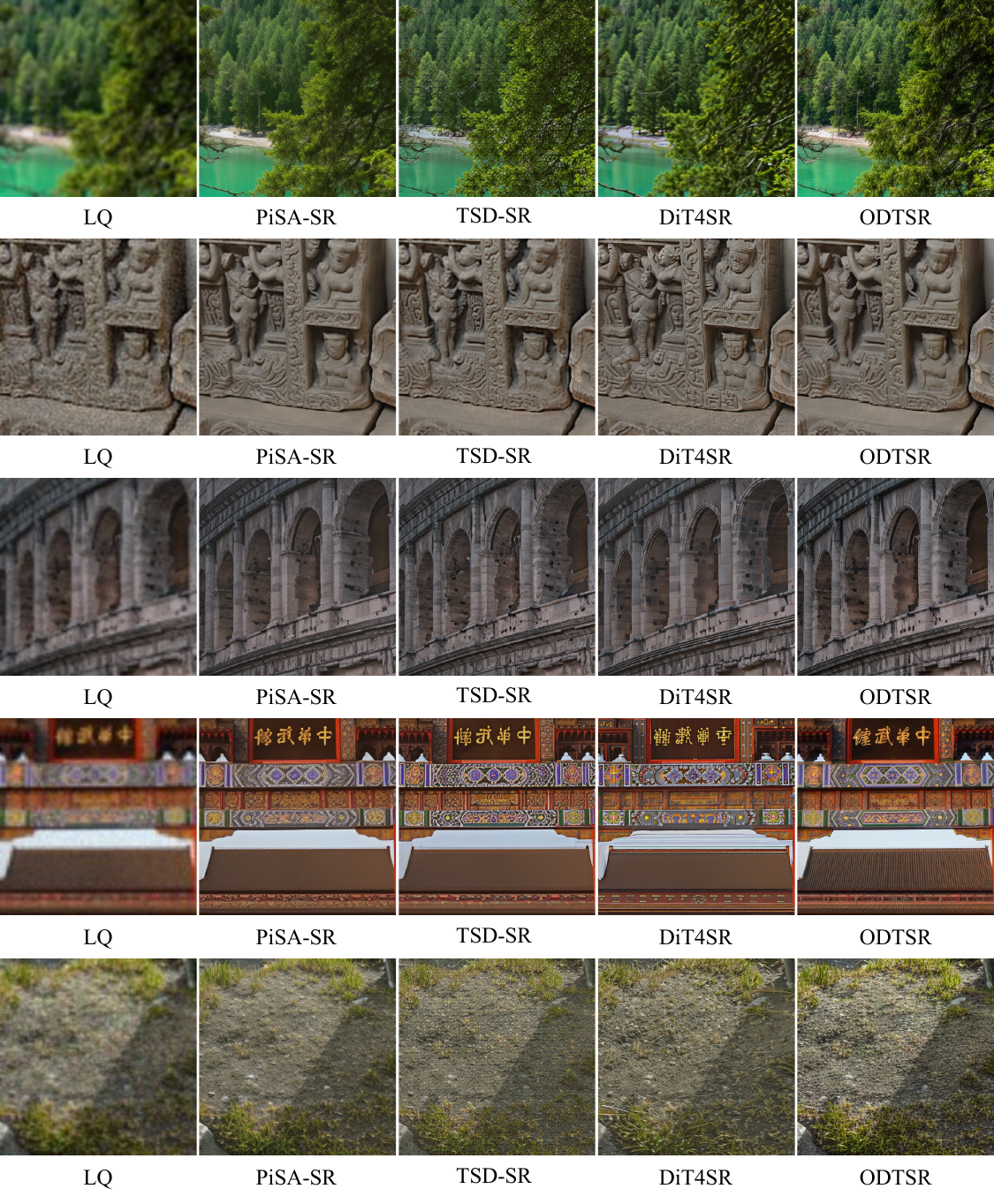}
    \caption{More qualitative results on \textbf{Real-ISR}. \textit{Fidelity Weight} is set to 1.0 and input prompt is empty in ODTSR. ODTSR achieves remarkable performance in fidelity and detail quality.}
    \label{fig:supp_realisr_1}
\end{figure*}

\begin{figure*}[p]
    \centering
    \includegraphics[width=\linewidth]{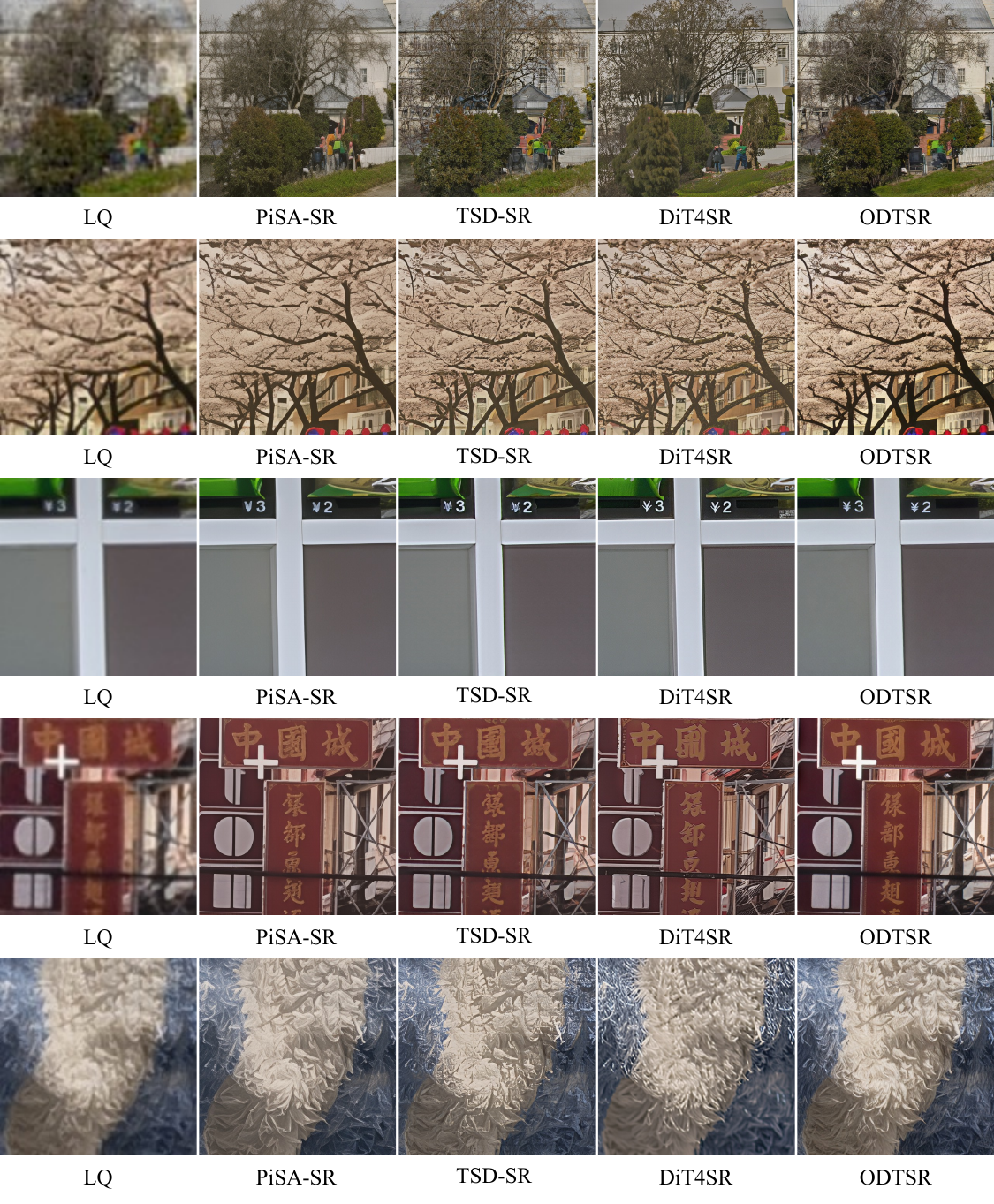}
    \caption{More qualitative results of \textbf{Real-ISR}. \textit{Fidelity Weight} is set to 1.0 and input prompt is empty in ODTSR. ODTSR achieves remarkable performance in fidelity and detail quality.}
    \label{fig:supp_realisr_2}
\end{figure*}

We present more Real-ISR results in \cref{fig:supp_realisr_1} and \cref{fig:supp_realisr_2}, comparing our ODTSR with three state-of-the-art methods PiSA-SR~\cite{sun2025pisasr} (one-step, based on Diffusion U-Net), TSD-SR~\cite{dong2025tsdsr} (one-step, based on Diffusion Transformer) and DiT4SR~\cite{duan2025dit4sr} (multi-step, based on Diffusion Transformer), which is identical to the settings in the user study. In generic Real-ISR setting, ODTSR achieves remarkable performance in fidelity and detail quality. For example, Row 2 of \cref{fig:supp_realisr_1} shows restored sculpture image with refined and realistic facial details and Row 5 of \cref{fig:supp_realisr_2} shows restored correct texture of the blanket with no prompt guidance.

\subsection{Controllable Real-ISR}
\begin{figure*}[p]
    \centering
    \includegraphics[width=\linewidth]{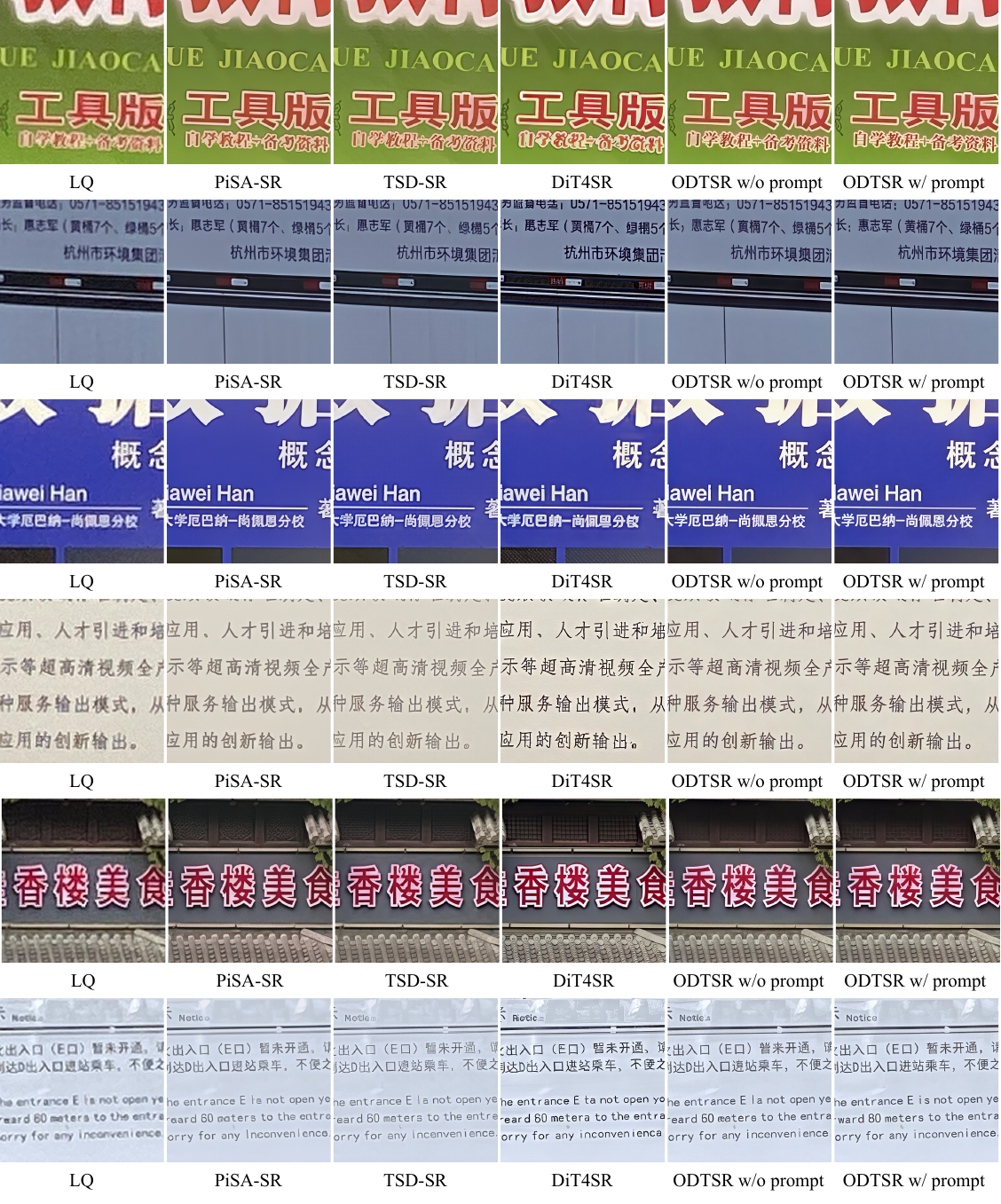}
    \caption{More qualitative results of \textbf{Controllable Real-ISR} on \textit{RealCE-val} dataset. \textit{Fidelity Weight} is set to 1.0 and we show results of ODTSR both with and without prompt (text annotation). ODTSR achieves higher restoration quality than other methods.}
    \label{fig:supp_realce_512}
\end{figure*}

\paragraph{Text scene} We present more results on RealCE-val, a scene text image super-resolution dataset. We show results of ODTSR both with and without prompt. As shown in \cref{fig:supp_realce_512}, ODTSR achieves higher restoration quality than other methods without prompt. When text annotation is incorporated as a prompt, some distorted characters were corrected while also demonstrating prompt controllability.

\begin{figure*}[p]
    \centering
    \includegraphics[width=\linewidth]{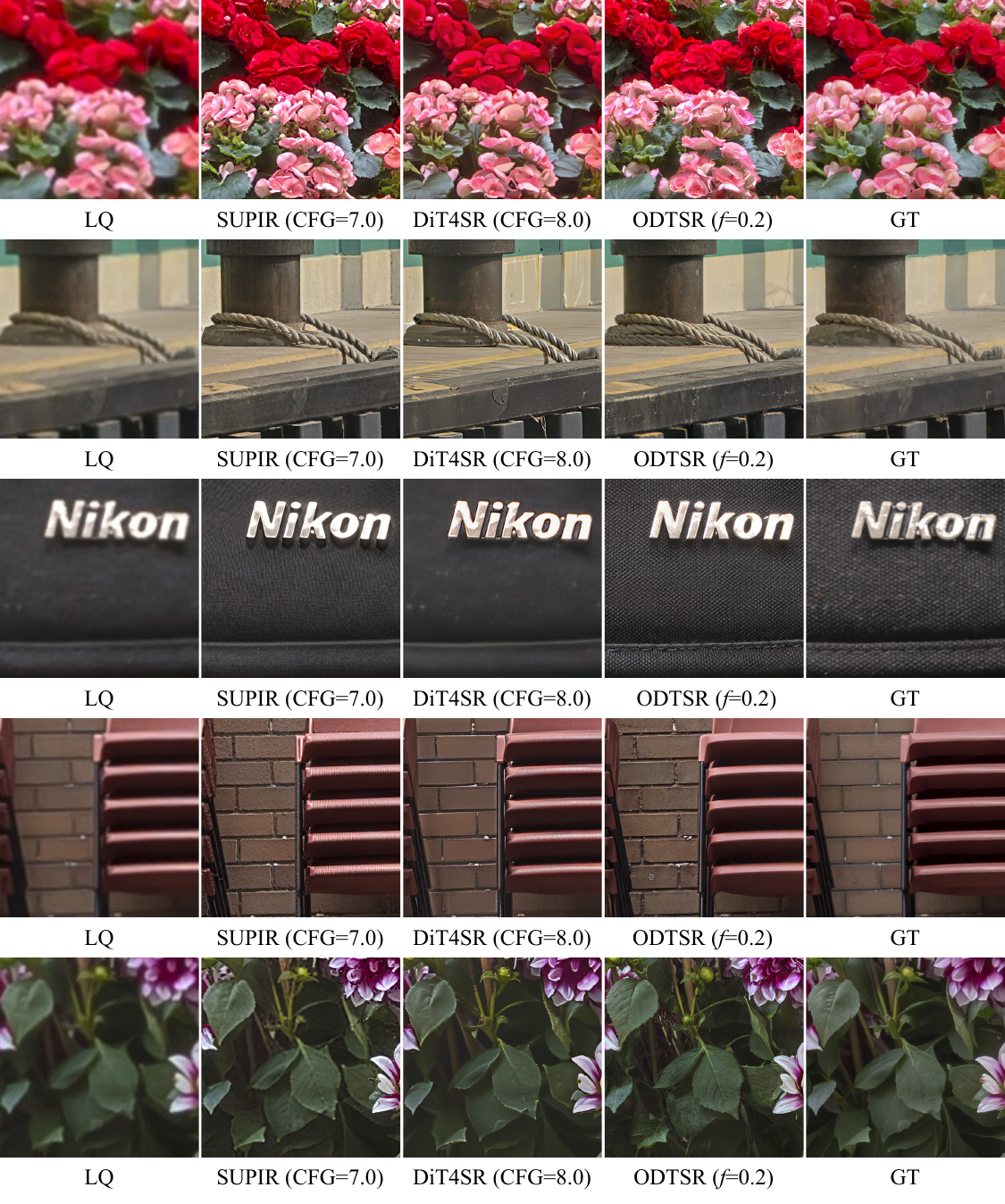}
    \caption{More qualitative results of \textbf{Controllable Real-ISR} on \textit{RealSR} dataset. \textit{Fidelity Weight} is denoted as \textit{f} and the input prompt is extracted from GT same across three methods. ODTSR restores more realistic and higher-quality details under prompt control.}
    \label{fig:supp_realsr_gt_prompt}
\end{figure*}

\paragraph{General scene} We present more results on RealSR with prompt extracted from GT for easier visual comparison. \cref{fig:supp_realsr_gt_prompt} shows that ODTSR, compared with multi-step methods SUPIR~\cite{yu2024supir} and DiT4SR~\cite{duan2025dit4sr}, restores realistic and high-quality details under prompt control. For example, in Row 4 of \cref{fig:supp_realsr_gt_prompt}, ODTSR restores correct details of seats and wall.

\begin{figure*}[p]
    \centering
    \includegraphics[width=\linewidth]{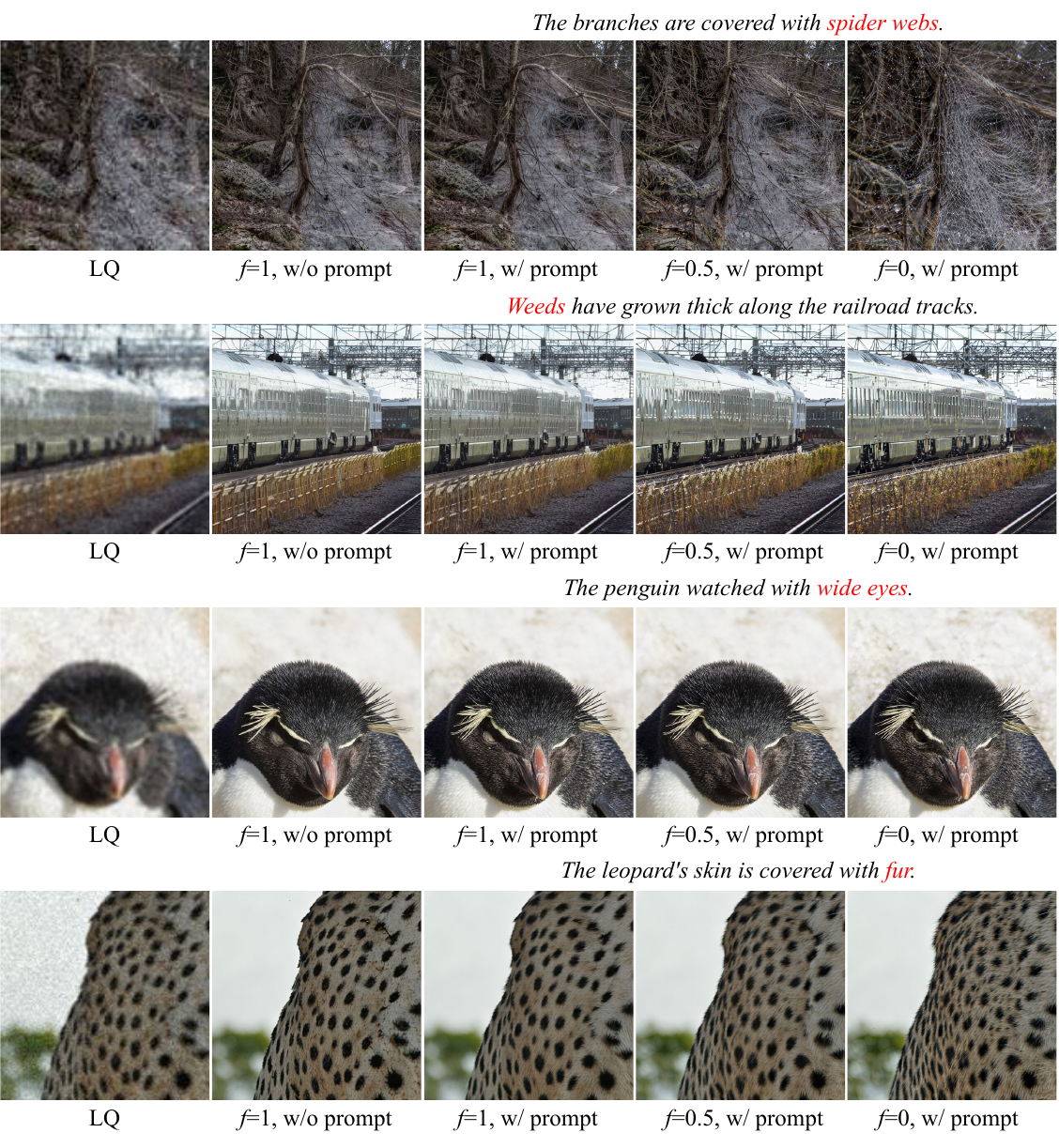}
    \caption{More qualitative results of \textbf{Controllable Real-ISR} with prompt and adjustable \textit{Fidelity Weight} (denoted as \textit{f}) on \textit{Div2k-val} dataset. As \textit{f} decreases from 1 to 0, detail generation and prompt adherence gradually strengthen, demonstrating controllability of ODTSR. The prompt corresponding to the enhanced details is marked in \textcolor{red}{red}.}
    \label{fig:supp_realsr_fidelity}
\end{figure*}

\paragraph{Adjustable Fidelity Weight} In \cref{fig:supp_realsr_fidelity}, we present results with adjustable \textit{Fidelity Weight} (denoted as \textit{f}) in ODTSR to demonstrate the scope and effectiveness of controllability. In general, as \textit{f} decreases from 1 to 0, detail generation and prompt adherence gradually strengthen.

\section{Limitation and Future Works}
Although ODTSR achieves high fidelity and prompt controllability, its large number of parameters leads to the high computational cost. We plan to apply various model acceleration techniques to mitigate computational costs. Additionally, the \textit{Fidelity Weight} in ODTSR is currently adjusted for the whole image, leaving room for more fine-grained control. We plan to explore more efficient and precise control approaches in the future.

\end{document}